# Development and Validation of a Large Language Model for Generating Fully-Structured Radiology Reports

Chuang Niu[1†] (PhD), Md Sayed Tanveer[1†], Md Zabirul Islam[1†], Parisa Kaviani[2] (MD), Qing Lyu[3] (PhD), Mannudeep K. Kalra[2] (MD), Christopher T. Whitlow[3] (MD), Ge Wang[1] (PhD)

1. Department of Biomedical Engineering, School of Engineering, Center for Computational Innovations, Center for Biotechnology & Interdisciplinary Studies, Rensselaer Polytechnic Institute, 110 8th Street, Troy, 12180, NY, USA.

2. Department of Radiology, Massachusetts General Hospital, Harvard Medical School, White 270-E, 55 Fruit Street, Boston, 02114, MA, USA.

3. Department of Radiology, Wake Forest University School of Medicine, Winston-Salem, 27103, NC, USA.

[†] Chuang Niu, Md Sayed Tanveer, and Md Zabirul Islam contribute equally.

## ABSTRACT

### Background

Current LLMs for creating fully-structured reports face the challenges of formatting errors, content hallucinations, privacy leakage issues when uploading data to external servers, and high computational cost with heavy prompting.

### Purpose

To develop a dynamic template-constrained decoding method and an open-source smaller LLM for accurately creating fully-structured and standardized LCS reports from varying free-text reports across institutions and demonstrate its utility in automatic statistical analysis and individual lung nodule retrieval.

### Materials and Methods

To develop dynamic template-constrained decoding method, 5,442 de-identified LDCT LCS radiology reports were included from two institutions in a retrospective study with IRB approvals. We constructed two evaluation datasets by labeling 500 pairs of free-text and fully-structured radiology reports and one large-scale consecutive dataset from January 2021 to December 2023. Two radiologists created a standardized template for recording 27 lung nodule features on LCS. We designed a dynamic-template-constrained decoding method to enhance existing LLMs for creating fully-structured reports from free-text radiology reports. Using consecutive structured reports, we automated descriptive statistical analyses and a nodule retrieval prototype.

Furthermore, to finetune a smaller LLM for fully-structured radiology reporting, we collected de-identified Chest CT reports without contrast. Based on this dataset, we then built an LLM-based workflow capable of creating fully-structured templates with 195 abnormality features and synthesizing paired free-text and fully-structured reports without any human annotations. Finally, we got a training dataset of 189,950 paired reports and a testing dataset of 499,65 paired reports.

The LLM performance was evaluated with F1 score, intersection over union (IoU), confidence interval, McNemar test, and z-tests.

**Results**

The best LLM with our inference method achieved high performance in creating fully-structured reports on cross-institutional datasets (F1 score: 0.976; 95% CI: 0.965-0.986 and F1 score: 0.969; 95% CI: 0.958-0.980), with neither formatting errors nor content hallucinations. Our method consistently improved the best open-source LLMs by up to 10.42% (97.60% vs 87.18%; p-value < 0.01), and outperformed GPT-4o by 17.19% (97.60% vs 80.41%; p-value < 0.01). The automatically derived statistical distributions were consistent with prior findings regarding attenuation, location, size, stability, and Lung-RADS. The retrieval system with structured reports allowed flexible nodule-level search and complex statistical analysis. Our developed software is publicly available for local deployment and further research. Our finetuned LLM with 8 billion parameters significantly improved the baseline model in terms of both IoU (0.183 vs 0.840) and F1 score (0.350 vs 0.899), and significantly reduced the inference time from 9 hours to 2.5 hours in processing 5,000 reports under the same settings.

**Conclusion**

Our LLM inference method and synthetic data fine-tuning approach accurately created fully-structured reports from free-text LCS radiology reports across institutions, defining state-of-the-art performance. Its utility was demonstrated in automatic statistical analysis and individual lung nodule retrieval.

**Abbreviations**

LLM = Large Language Model, FSR = Fully-Structured Report, LSR = Loosely-Structured Report, LCS = Lung Cancer Screening, LDCT = Low-Dose CT, GGN = Ground Glass Nodule, JSON = JavaScript Object Notation, CI = Confidence Interval, Lung-RADS = Lung CT Screening Reporting and Data System

# 1. Introduction

Structured and standardized reporting in radiology has been recognized to improve quality, consistency, and actionability of radiology reports, leading to optimized clinical workflow and patient outcomes [1-4]. Structured reports can be divided into two levels: structured layout (level-1) and structured content (level-2) [5]. We refer the level-1 reports to loosely-structured reports (LSR) that organizes free-text findings with subheadings in a specific order, such as *lung nodules, lungs, pleura, heart, etc.*, in LCS. Most radiology reporting involves free-text contents in either unstructured or loosely-structured layout. Similarly, we define fully-structured reports (FSR) as those with the structured contents consisting of comprehensive predefined features, such as nodule attenuation, location, margin, shape, size, stability, etc., for describing lung nodule. In FSR, a set of standardized candidate values are predefined for each feature.

Radiology reports contain rich information for research and clinical purposes. To effectively perform data mining, FSR provides standardized, discrete data elements that are easy to search and analyze [6]. Despite its various benefits, FSR is rare in clinical routine [7]. In the context of rising imaging volume, size of individual imaging exams (particularly the cross-sectional techniques), workforce shortage, and emphasis on short radiology report turnaround time, current software with a “click-heavy” process becomes a major obstacle to FSR adoption at the cost of reduced efficiency and distraction from accurate reporting [8].

Advancement in LLMs [9] provides unprecedented opportunities to create FSR seamlessly in the radiological workflow. For example, GPT-4 was leveraged to convert 170 free-text CT and MRI reports to structured reports at level-1 [10]. In a most recent study [11], GPT-3.5 and GPT-4 were prompted to create synoptic reports at level-1 from 180 original CT reports for pancreatic ductal adenocarcinoma, where surgeons demonstrated greater accuracy in categorizing respectability using the converted reports than original ones. However, there has been no study so far demonstrating the ability of LLMs for the desired FSR. Moreover, current studies [10-13] used the proprietary LLMs which require uploading radiology reports to the cloud server, raising privacy concerns especially for large-scale studies [14]. This concern can be addressed by deploying the latest and similarly powerful open-source Llama 3.1 with 405 billion parameters on the local server [15]. However, there are two technical challenges when applying an LLM for structured reporting: formatting errors and content hallucinations. JSON is used as a standard format to output the structured contents by LLM, but LLMs cannot ensure the generated JSON format always correct, leading to failed conversions. On the other hand, the well-known hallucinations of LLMs cannot be tolerated in the healthcare applications [16, 17]. These challenges must be overcome to establish an accurate and reliable LLM-based fully-structured reporting. Additionally, while many initiatives were launched to promote structured reporting, a cross-institutional study is lacking [7].

Although open-source LLM have demonstrated strong performance in generating structured radiology reports, their deployment is limited by high computational cost and long inference time due to the long instructions prompt in the fully-structured reports. Running these models on large-scale datasets is impractical for most research and clinical environments. On the other hand, smaller models such as Llama 3.1–8B are more lightweight but still inefficient in their base form. Because they do not inherently understand the structured reporting schema, they also require long prompts that include detailed templates and instructions.

The purpose of this study is to develop a dynamic template-constrained decoding method to empower the open-source LLMs and fine-tune an smaller LLM based on a synthetic data curation workflow for automatically generating FSR from free-text contents. We will define a cross-institutional template, curate cross-institutional evaluation datasets, and demonstrate the actionability of FSR through automatic statistical analysis, nodule-level information retrieval on a large-scale consecutive dataset, and extensive analysis on the complete chest CT reports conversion.

# 2. Materials

With IRB approvals, the deidentified LDCT LCS radiology reports with the associated age and sex information were collected from two quaternary hospitals (denoted by Institution-1 and Insitutiona-2) in this retrospective and cross-institutional study in compliance with the Health Insurance Portability and Accountability Act. For LDCT LCS, both institutions used the LSR but had different subheadings. Specifically, the Institution-1 reports usually include *lung nodules, COPD, pleura, heart, coronary artery calcification, mediastinum/hilum/axilla, and other findings* to present findings, while institution-2 usually uses *lines/tubes/devices, lungs, pleura, mediastinum/heart, chest wall, soft tissues, upper abdomen, and bones* to represent findings. Following each subheading, various styles of free-text descriptions were written by radiologists.

**Table 1: Structured template for lung nodule reporting.**

| Features | Candidate Set |
|---|---|
| Nodule ID | Integer numbers: 1-50, null |
| Series ID | Integer numbers: 1-2000, null |
| Image ID | Integer numbers: 1-5000, null |
| Lobe | (1) right upper lobe, (2) right middle lobe, (3) right lower lobe, (4) left upper lobe, (5) lingula, (6) right lower lobe, (7) null |
| Segment | (1) apical, (2) posterior, (3) anterior, (4) lateral, (5) medial, (6) superior, (7) medial basal, (8) anterior basal, (9) lateral basal, (10) posterior basal, (11) apico-posterior, (12) anteromedial basal, (13) inferior, (14) null |
| Fissure | (1) right minor fissure, (2) right major fissure, (3) left major fissure, (4) right fissural, (5) left fissural, (6) fissural, (7) right perifissural, (8) left perifissural, (9) perifissural, (10) null |
| Peripheral | (1) subpleural, (2) pleural-based, (3) peripheral, (4) null |
| Tracheobronchial | (1) airway, (2) tracheal, (3) endotracheal, (4) endobronchial, (5) peribronchial, (6) peribronchovascular, (7) bronchocentric, (8) bronchovascular, (9) null |
| Perihilar | (1) perihilar, (2) null |
| Lung | (1) left, (2) right, (3) bilateral, (4) null |
| Type | (1) solid, (2) part-solid, (3) mixed attenuation, (4) cystic, (5) ground glass, (6) hazy, (7) nonsolid, (8) fluid/water, (9) fat, (10) calcified, (11) part-calcified, （12）noncalcified, (13) cavitary, (14) null |
| Calcification Patterns | (1) diffuse, (2) central, (3) lamellated, (4) popcorn, (5) eccentric, (6) dense, (7) dendriform, (8) punctate, (9) linear, (10) null |
| Margin | (1) spiculated, (2) smooth, (3) lobulated, (4) fuzzy, (5) irregular, (6) null |
| Shape | (1) oval, (2) lentiform, (3) triangular, (4) round, (5) bilobed, (6) rectangular, (7) polygonal, (8) spherical, (9) irregular, (10) ovoid, (11) tubular, (12) branching, (13) null |
| Long Axis (mm) | Float numbers: 0.00-200.00, null |
| Short Axis (mm) | Float numbers: 0.00-200.00, null |
| Third Axis (mm) | Float numbers: 0.00-200.00, null |
| Average Diameter (mm) | Float numbers: 0.00-200.00, null |
| Part-Solid Diameter (mm) | Float numbers: 0.00-200.00, null |
| Volume ($mm^3$) | Float numbers: 0.00-10000.00, null |

| Mass (mg) | Float numbers: 0.00-10000.00, null |
|---|---|
| Qualitative Size | (1) small, (2) tiny, (3) micronodule, (4) punctate, (5) subcentimeter, (6) large, (7) null |
| Stability | (1) stable, (2) increase, (3) decrease, (4) new, (5) changed, (6) baseline, (7) interval development,<br>(8) resolved, (9) null |
| Lung-RADS | (1) 0, (2) 1, (3) 2, (4) 3, (5) 4A, (6) 4B, (7) 4X, (8) null |
| Overall Lung-RADS | (1) 0, (2) 1, (3) 2, (4) 3, (5) 4A, (6) 4B, (7) 4X, (8) null |
| Recommend Imaging | (1) LDCT, (2) LDCT, PET/CT, (3) LDCT, Tissue sampling, (4) LDCT, PET/CT, Tissue sampling, (5) LDCT, Diagnostic CT, (6) LDCT, Diagnostic CT, PET/CT, (7) LDCT, Diagnostic CT, PET/CT, Tissue sampling, (8) Diagnostic CT, (9) Diagnostic CT with contrast, (10) Diagnostic CT, PET/CT, (11) Diagnostic CT with contrast, PET/CT, (12) Diagnostic CT, PET/CT, Tissue sampling, (13) Diagnostic CT with contrast, PET/CT, Tissue sampling, (14) PET/CT, (15) PET/CT, Tissue sampling, (16) Tissue sampling, (17) null |
| Imaging Interval | (1) 1 month, (2) 1-2 months, (3) 1-3 months, (4) 2-3 months, (5) 3 months, (6) 3-6 months, (7) 4-6 months, (8) 6 months, (9) 6-12 months, (10) 12 months (11) null |
| Number of Nodules | Integer numbers: 0-50, null |

Note.— Lung-RADS = Lung CT Screening Reporting and Data System

To study automatic creation of the FSR from LSR with LLM, two radiologists first drafted a detailed, standardized template (Table 1) with the purpose of comprehensive, all-feature lung nodule reporting. The template consists of 28 features, including 24 nodule-level features that are grouped together as a nodule descriptor to describe an individual nodule, 3 features to present patient management at report-level, and 1 auxiliary feature for the number of nodule descriptors to cover all individual lung nodules in the original report. Each feature has a set of standardized terms as its candidate values. A single nodule descriptor may present multiple nodules, e.g., when the number of descriptors cannot be exactly determined from the description, such as "*small bilateral calcified nodules*", then a single descriptor is created to represent these nodules.

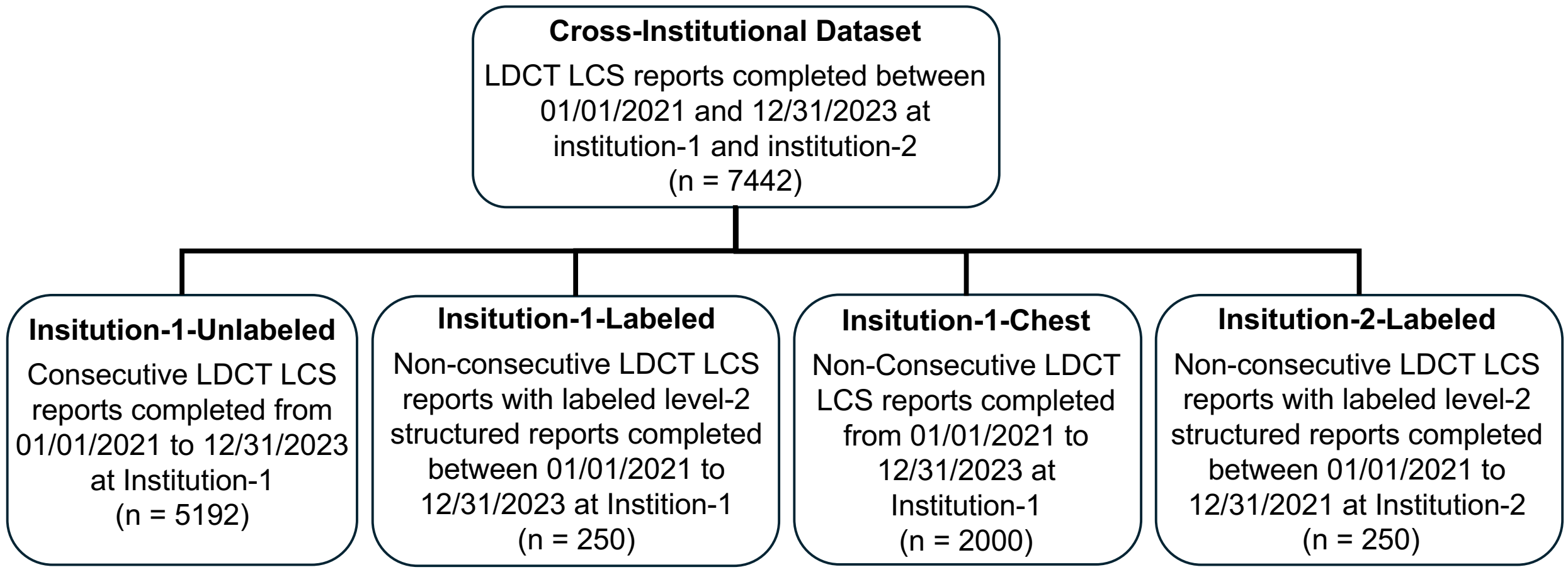

Figure 1. Cross-institutional dataset construction. LDCT = low dose computed tomography. LCS = lung cancer screening.

**Table 2: Distributions of cross-institutional datasets by age and sex.**

| Age | <55 y | | 55-65 y | | 65-75 y | | ≥75 y | |
|---|---|---|---|---|---|---|---|---|
| Sex | Male | Female | Male | Female | Male | Female | Male | Female |
| Insitution-1-Labeled | 4 | 3 | 58 | 56 | 64 | 47 | 9 | 9 |
| Institution-2-Labeled | 7 | 6 | 62 | 64 | 50 | 44 | 8 | 9 |
| Insitution-1-Unlabeled | 37 | 72 | 1122 | 1162 | 1185 | 1200 | 205 | 209 |

Note.— Insitution-1-Labeled and Institution-2-Labeled mean the datasets from the first and the second institutions respectively. Each of the datasets contains 250 pairs of original and labeled structured reports. Insitution-1 is the dataset including 5,192 original reports from the first institution from January 1, 2021 to December 31, 2023.

Our cross-institutional dataset construction is summarized in Figure 1. To quantitatively evaluate LLM performance, we built two evaluation datasets by randomly selecting 500 LSRs from the two institutions and manually converting them to the FSRs according to the predefined template. For automatic statistical analysis and individual lung nodule retrieval, we collected a large-scale dataset that consists of all the 5,192 consecutive LDCT LCS reports from January 1, 2021, to December 31, 2023 at Institution-1. The report distributions by age and sex of the three datasets are summarized in Table 2. To fine-tuning an LLM for generating fully-structured chest CT reports including as many chest abnormalities as possible, we also collected 2,000 reports from chest CT without contrast exams between January 1, 2021, to December 31, 2023 at Institution-1.

# 3 Dynamic Template-Constrained Decoding Method

The existing studies focus on designing the system instruction to prompt LLMs for radiology applications [12]. However, such prompting cannot eliminate either formatting errors or content hallucinations in creating FSR with LLMs. To overcome these challenges, we propose a dynamic-template-constrained decoding method. The basic idea is to use a structured and standardized template to strictly constrain the LLM output. As shown in Figure 2, our LLM inference method takes the system instruction and the conventional LSR as input, uses the structured and standardized template to strictly constrain every output token during decoding, and then produces the FSR accordingly.

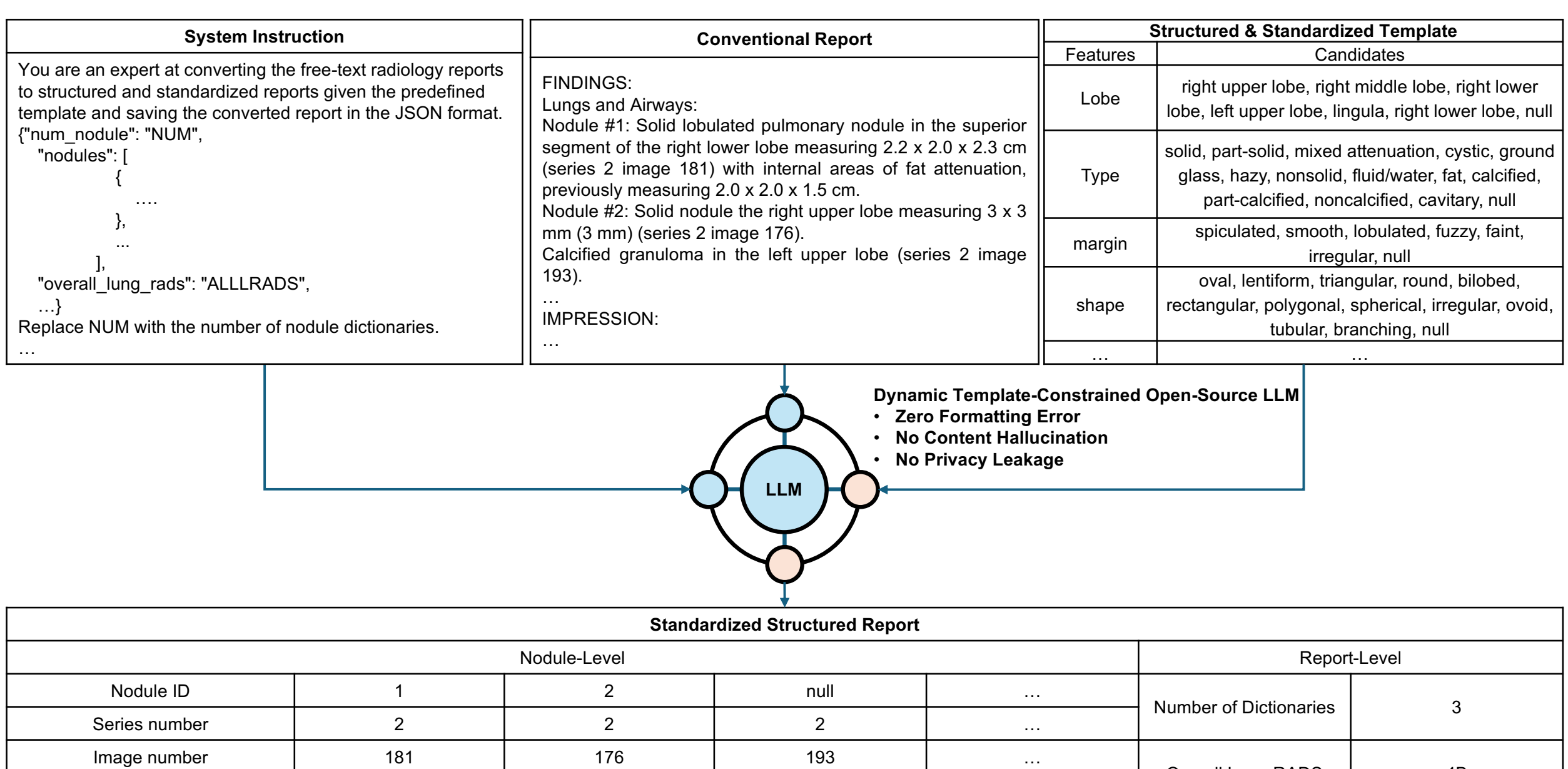

| Standardized Structured Report | | | | | | |
|---|---|---|---|---|---|---|
| Nodule-Level | | | | | Report-Level | |
| Nodule ID | 1 | 2 | null | … | Number of Dictionaries | 3 |
| Series number | 2 | 2 | 2 | … | | |
| Image number | 181 | 176 | 193 | … | Overall Lung-RADS | 4B |
| Lobe | RLL | RUL | LUL | … | | |
| Segment | superior | null | null | … | Recommend Imaging | PET/CT |
| Type | solid | solid | calcified | … | | |
| Long axis | 23 | 3 | null | … | | |
| Short axis | 20 | 3 | null | … | Imaging Interval | 1 month |
| Third axis | 22 | null | null | … | | |
| Average diameter | 21.67 | 3 | null | … | … | … |
| Status | increase | null | null | … | | |
| Lung-RADS | 4B | null | null | … | | |
| … | … | … | … | … | | |

Figure 2. Framework of the dynamic-template-constrained LLM for fully-structured reporting.

Our designed dynamic-template-constrained LLM is shown in Figure 3. We modified the current LLM inference process in Figure 3 (a) and built a novel dynamic-template-constrained inference architecture in Figure 3 (b), with the differences highlighted in orange. To handle varying numbers of nodules in different reports, a dynamic template is designed to dynamically determine and create the required number of nodule descriptors during inference as shown in Figure 3 (c). Our decoding process in Figure 3 (d) illustrates that the LLM is constrained to exactly output the predefined format text and select one of the predefined candidate values for each feature, ensuring neither JSON formatting errors nor content hallucinations. Our dynamic-template-constrained decoding method is detailed in Appendix.

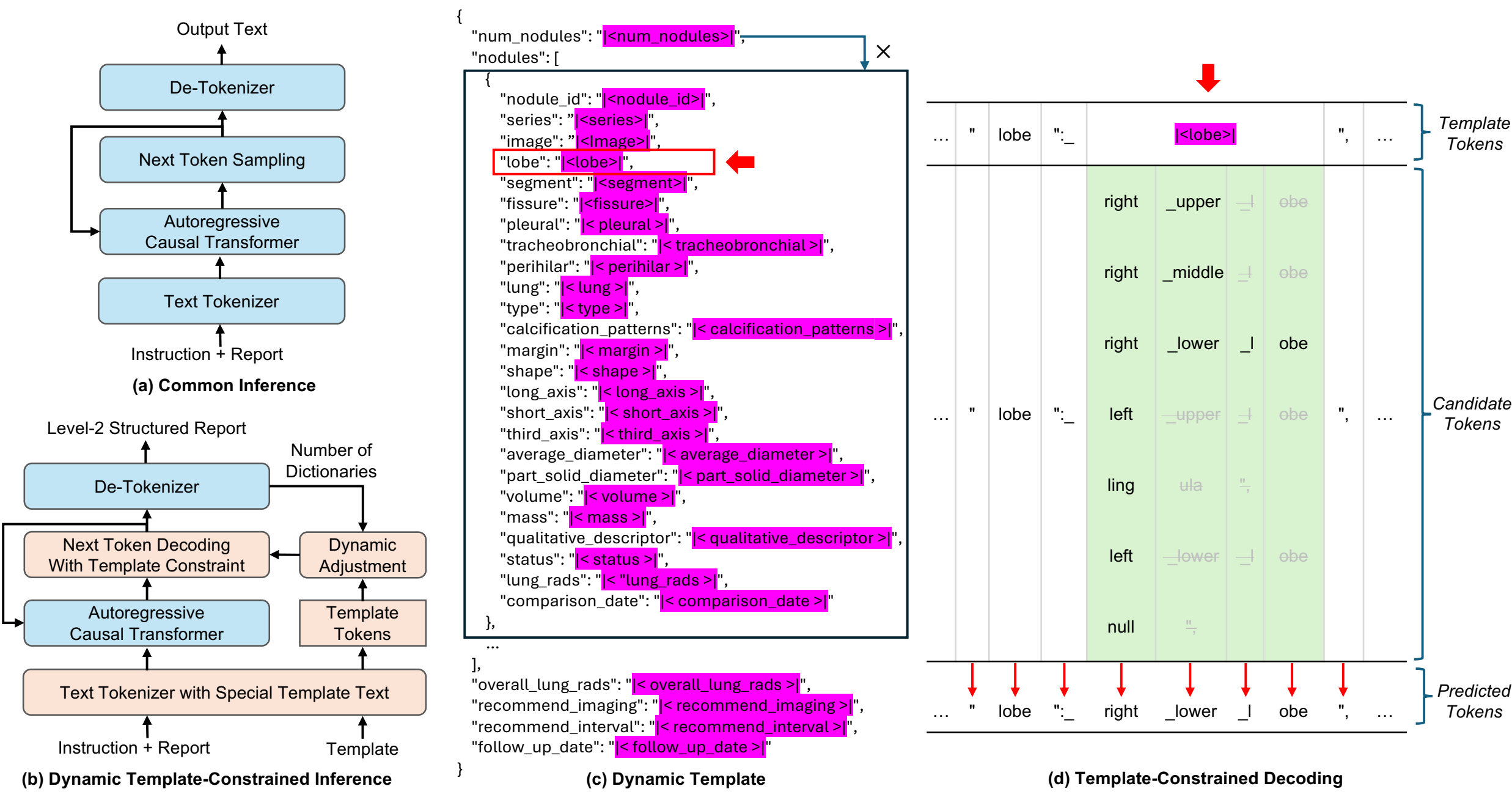

Figure 3. Dynamic template-constrained decoding. (a) The inference architecture of current LLMs, (b) our dynamic template-constrained inference architecture, (c) dynamic JSON template including template text without highlighted and special template text highlighted in pink, and (d) illustration of dynamic template-decoding process, where the candidate text corresponding to the special template text |<*lobe*>| are highlighted in green.

Our dynamic template-constrained method was developed based on the vLLM project [18], which is an open-source library for LLM inference and serving. In principle, our approach can be applied to enhancing any LLM while its implementation needs to modify the common inference architecture. We studied its superiority on the most powerful open-source LLMs, including Llama-3.1 (8B, 70B, 405B) [15], Qwen-2 (72B) [19], and Mistral-Large (123B) [20]. All these open-sourced models were downloaded from Hugging Face with approval. We also evaluated GPT-4o on the Insitution-1-Labeled dataset for comparison using the same system instruction. Due to the institutional policy, the other two datasets in Table 2 are not allowed to be uploaded to any external sever to use proprietary LLMs. Since the source codes of GPT-4o is not publicly available, we cannot implement our method to enhance it. The software development and experiments were conducted on a local server with 8xH100 GPUs. To facilitate further research of LLMs in structured radiology reporting, our vLLM-Structure software, example data, and tutorials have been made publicly available at https://github.com/niuchuangnn/vllm_structure.

Table 3. Number of samples generated for finetuning LLM-based parser/converter

| Regions | Training (Fine-tuning) | Testing (Inference) | Testing (Subsampled) |
|---|---|---|---|
| Chest wall/MSK | 30000 | 9100 | 1000 |
| Thoracic Inlet/Central Airways | 30300 | 10140 | 1000 |
| Heart/Vessels | 30060 | 9940 | 1000 |
| Mediastinum/Hila/Axilla | 30040 | 9960 | 1000 |
| Lungs/Pleura | 69550 | 10825 | 1000 |
| Total | 189950 | 49965 | 5000 |

# 4. LLM Finetuning

## 4.1 Data Curation and Synthetic Report Generation

Our primary objective was to evaluate the ability of language models to parse free-text radiology reports and generate structured tables, rather than to capture the full variability of natural language. Initially, we have access to approximately 2,000 de-identified CT reports; these real reports lacked corresponding structured templates required for supervised finetuning. Generating such templates from raw reports posed two major challenges: (i) automatic pipelines based on prompting existing LLMs were slow and computationally expensive at scale, and (ii) even when produced, the structured outputs were often inconsistent in format or contained incorrect values, undermining quantitative analysis and limiting their utility for training foundation models.

To overcome these limitations, we designed a synthetic data generation strategy that ensured structural consistency while still approximating the variability of real clinical language. The process involved three stages:

1. **Template-driven JSON generation**: Feature keys and values were randomly sampled from a predefined schema spanning multiple anatomical regions (thoracic inlet/central airways, chest wall/musculoskeletal system, mediastinum/hila/axilla, lungs/pleura, heart/vessels). The five templates were formed by analyzing the features appearing in the 2000 real CT reports. Each feature was assigned categorical values or numeric measurements within clinically realistic ranges, yielding ground truth structured tables.
2. **Free-text synthesis**: Each JSON table was transformed into a narrative description using rule-based templates. Controlled variation was introduced in terminology (e.g., “normal” vs. “not abnormal”), unit representation (cm vs. mm), and negation forms (“absent” vs. “not present”) to capture diverse clinical phrasing styles.
3. **Report integration**: The synthesized free-text sections were embedded into real and full free-text radiology reports, producing hybrid reports that preserved natural formatting while maintaining guaranteed alignment with the structured template. Out of 2,000, 1,500 real reports were used to fine-tune the LLM, and 500 were used for testing the performance. Each real report was injected with 20 different synthetic segments to form 20 times as many hybrid reports. This resulted in a training (fine-tuning) dataset of approximately 190,000 samples and a testing (inference) dataset of approximately 50,000 samples. However, we took a small, uniformly distributed subset of 5,000 samples to compare with the base (non-finetuned) model.

This approach yielded a large, paired dataset of free-text and structured representations suitable for supervised finetuning. The paired hybrid report - structured synthetic table report generation process can be found in Figure 4.

The number of samples for both fine-tuning and inference can be seen in Table 3. Feature-wise distributions for the different datasets are illustrated in Figure 5. From Table 3 and Figure 5, it can be seen that the feature appearances are not uniform, mainly because the structured template generation has a nested uniform sampling (feature number selection and then uniform feature pooling). The lung/pleura segment, in particular, has twice the number of samples compared to the other segments due to a greater importance on lung cancer and nodule screening. By training on this dataset, smaller LLMs internalized the reporting schema, enabling them to act as efficient free-text parsers and structured table generators. These fine-tuned models can then be deployed to accelerate the creation of large-scale structured templates from real-world clinical reports.

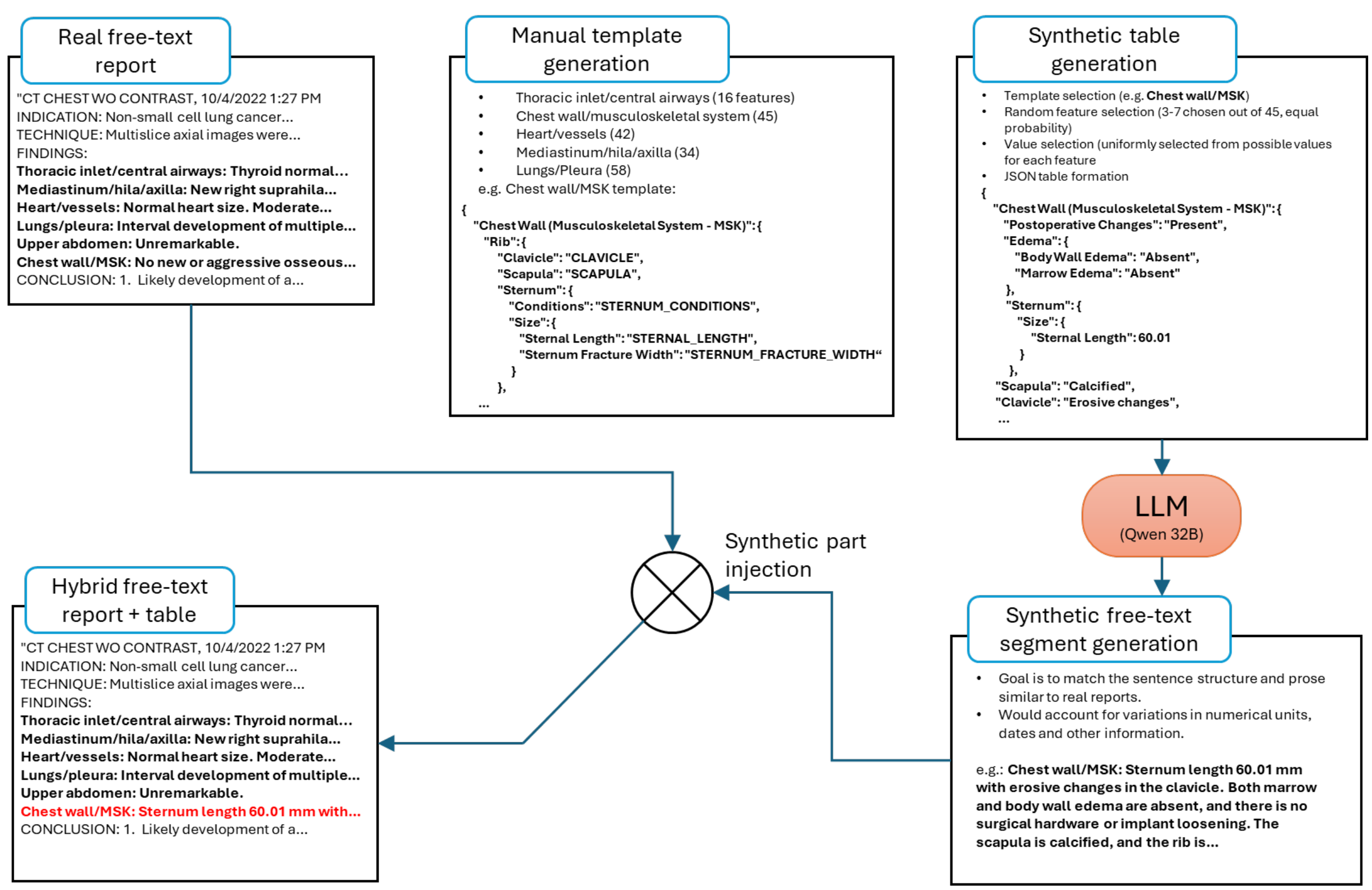

**Figure 4.** *Pipeline for synthesis report generation.* Here, an example of generating the hybrid full free-text report is shown. This way, it is possible to augment real CT scan reports to generate sufficient hybrid (real+synthetic) samples to finetune the LLM model.

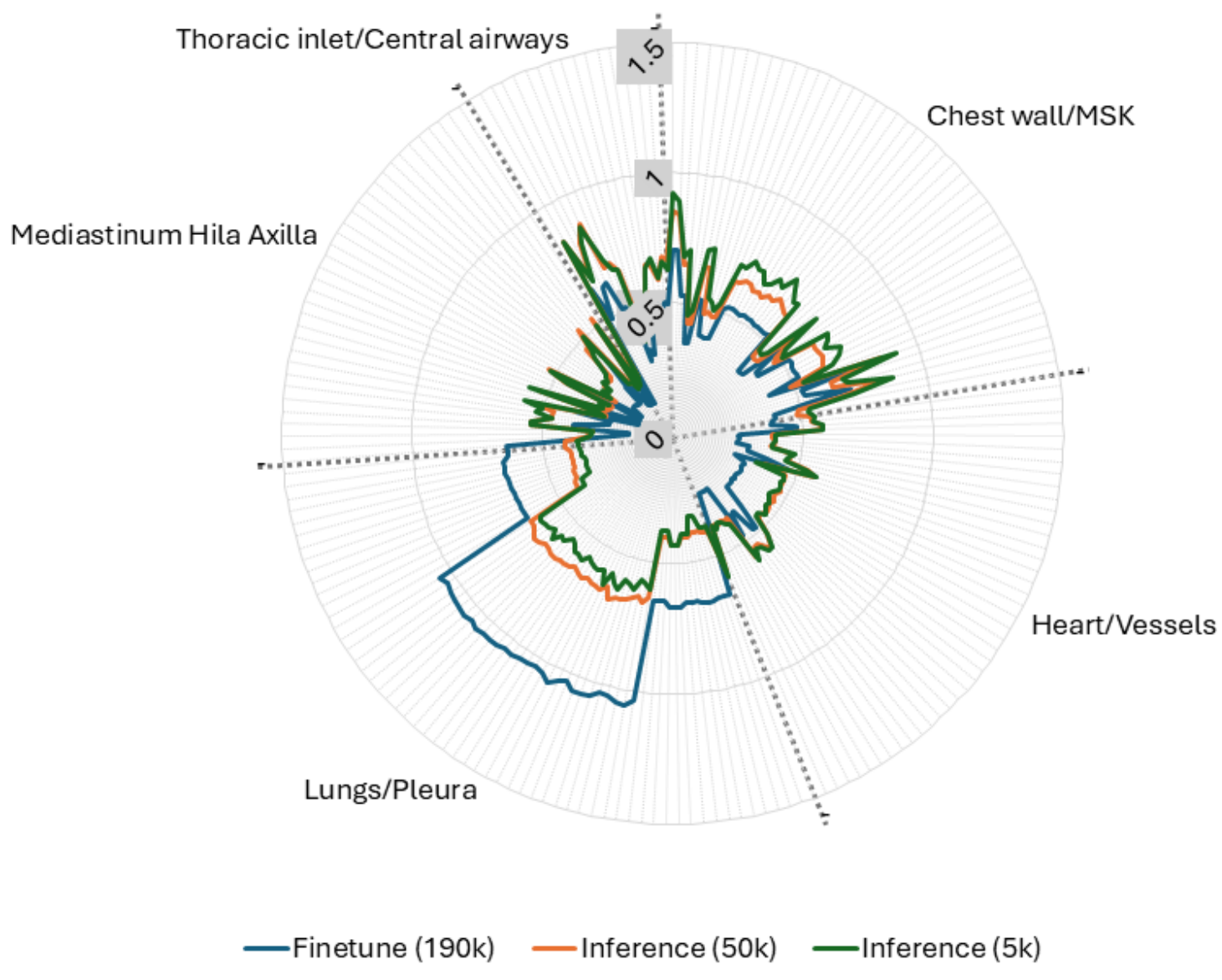

**Figure 5.** *Feature wise distribution of their appearance (%) in the different datasets.* Here, from the distribution, it is evident that the features across different datasets are fairly similarly distributed, owing to the similar synthetic data generation scheme, except for the lungs/pleura features in the training data, which have higher weights on the lung nodule-related features.

## 4.2 Model Training

We finetuned the Llama 3.1–8B Instruct model on the synthetic radiology dataset using the Torchtune framework in a distributed setting across two NVIDIA H100 GPUs, with mixed precision (bf16) enabled to maximize throughput and reduce memory overhead. Full parameter finetuning was employed with an effective batch size of 32, achieved through a per-device batch size of 4 and gradient accumulation steps of 8. Optimization was performed using AdamW with a learning rate of 2e-5, weight decay of 0.01, and gradient clipping set to 1.0. The training was carried out over five sequential runs, each corresponding to one epoch, for a total of five epochs. The tokenizer was configured with a maximum sequence length of 8,192 tokens to accommodate long free-text reports and structured outputs. To reduce GPU memory consumption, activation checkpointing was enabled, and sharded embeddings were applied to vocabulary-heavy layers for improved efficiency.

Training progress was logged with Weights & Biases, enabling close monitoring of throughput, learning rate dynamics, and loss convergence. Throughput remained stable across epochs, averaging between 700 and 800 tokens per second per GPU, with only occasional dips associated with checkpointing or system-level fluctuations. Learning rate schedules followed two distinct patterns: most runs used a fixed learning rate of 2e-5, while others incorporated cosine decay with warm-up, producing the expected linear ramp-up followed by gradual decay. Both strategies supported stable optimization without divergence. The training loss decreased steadily over the five epochs: early fluctuations in the first epoch diminished as training progressed, with convergence stabilizing around 0.02–0.03. These results confirmed that the model effectively internalized the structured reporting schema while avoiding overfitting to synthetic patterns.

### 4.3 Inference and Evaluation

We evaluated both the finetuned and base models using the LM-Eval harness, configured with a custom evaluation task designed for structured table generation. The evaluation dataset consisted of 5,000 hybrid radiology reports for direct comparison between models, and an extended evaluation of 49,965 reports was additionally conducted for the finetuned model to assess large-scale performance. Each input consisted of an instruction followed by a free-text report, and the model was required to generate a valid JSON object as output. Evaluation metrics included chrF, which provides a character-level similarity score, and Intersection-over-Union (IoU), which measures alignment of feature names and values within the generated structured tables.

The baseline Llama 3.1–8B Instruct model relied on long system prompts containing detailed templates and feature definitions, resulting in high token usage per sample. On average, input sequences reached 9,000–10,000 tokens, frequently approaching or exceeding the 8,192-token context limit. This led to left-truncated inputs, incomplete outputs, and extended inference times. Processing 5,000 reports required approximately 9 hours on two NVIDIA H100 GPUs with a batch size of 16.

In contrast, the finetuned Llama 3.1–8B model internalized the reporting schema during training, which eliminated the need for long prompts at inference. As a result, average token usage per sample was reduced to ~2,800–3500 tokens, well within the context window. The model consistently produced structured outputs capped at 1,024 tokens without truncation. Under identical hardware conditions, inference time for 5,000 reports decreased to 2.5 hours, representing more than a threefold speed-up. When scaled to the larger 49,965-report dataset, the model maintained both efficiency and structural consistency, demonstrating its practicality for large-scale deployment.

## 4.4 Automatic Structured Report Creation and Applications

All collected reports from the two institutions were automatically converted to the FSRs using our developed software. All open-source LLMs were implemented in both the original inference and the proposed dynamic template-constrained decoding method. The structured reports were saved in the JSON format. All models shared the same system instruction and structured template. We studied two use cases based on the created FSRs. First, the statistical distributions of various nodule features were automatically calculated by age and sex on the Insitution-1 dataset without any human annotations. Second, a nodule-level retrieval system was built. By inputting a combination of nodule features in the retrieval system, all the individual nodules met with request features will be listed, and the corresponding statistics and located CT images will be displayed.

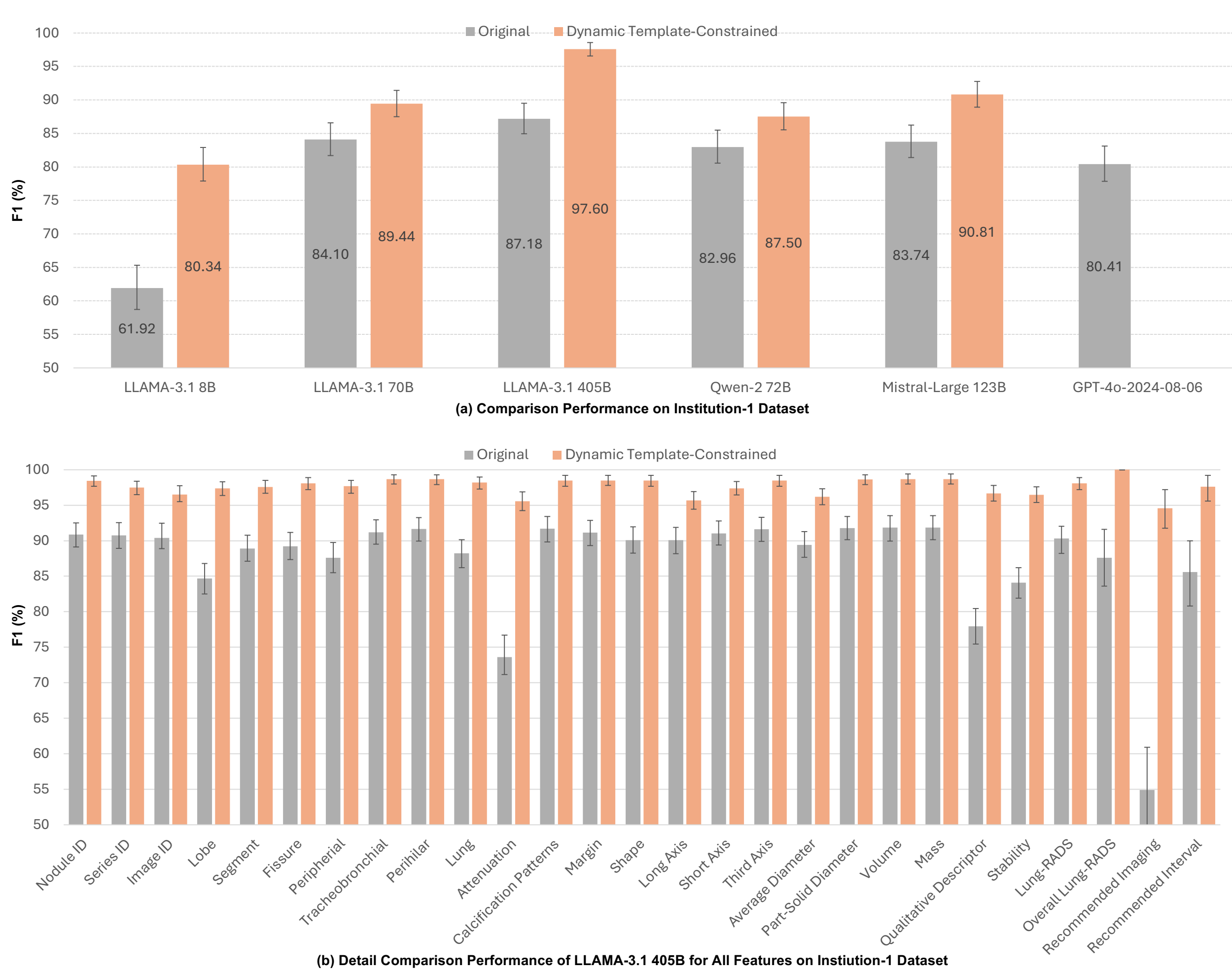

Figure 6. Comparative results of different LLMs on the Institution-1-Labeled dataset. (a) Average F1 scores and 95% confidence intervals of LLMs with and without our dynamic template-constrained encoding method are shown. (b) The F1 scores and 95% confidence intervals of the best LLAMA-3.1 model with 405 billion parameters for all features are shown.

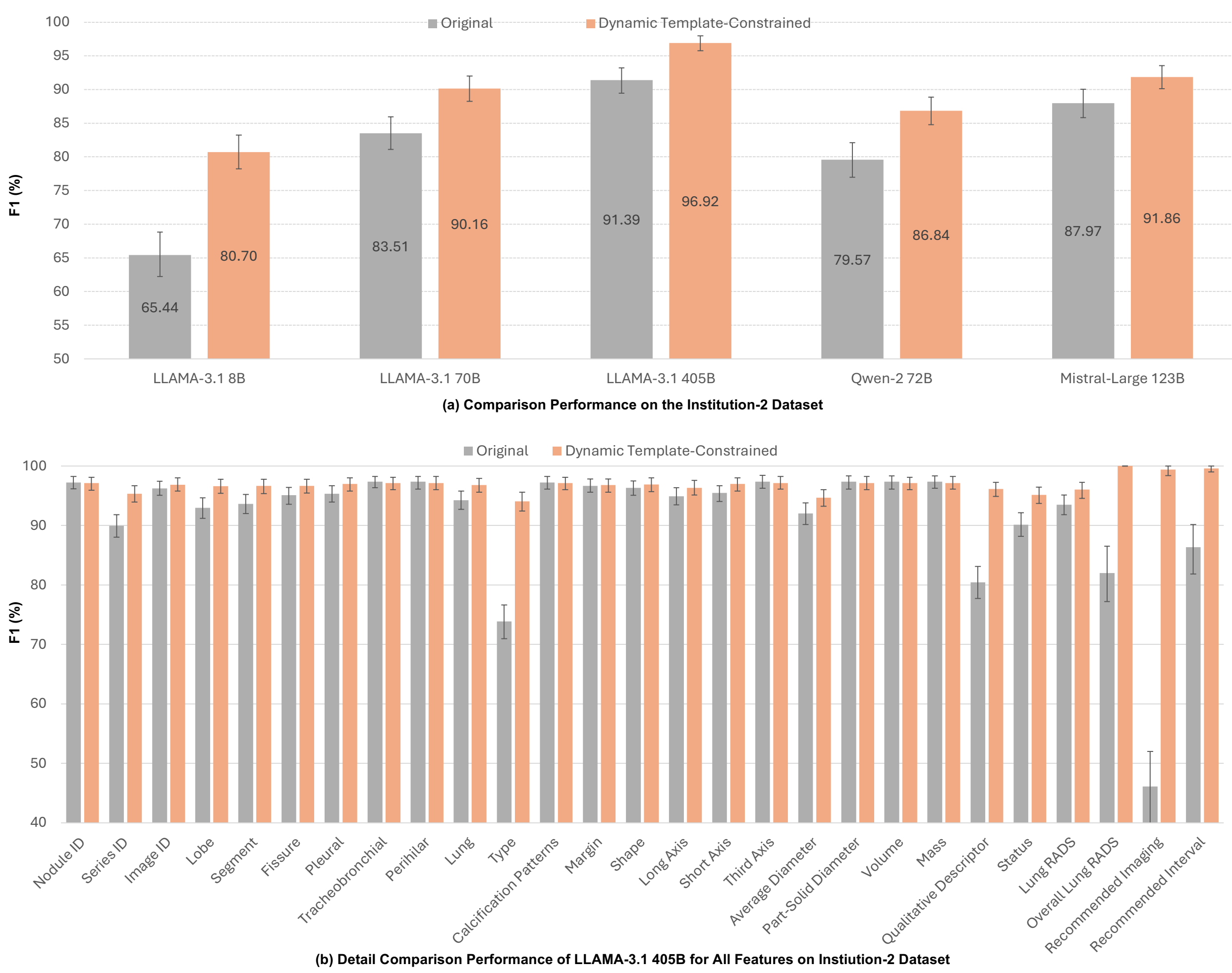

Figure 7. Comparative results of different LLMs on the Institution-2-Labeled dataset. (a) Average F1 scores and 95% confidence intervals of LLMs with and without our dynamic template-constrained encoding method are shown. (b) The F1 scores and 95% confidence intervals of the best LLAMA-3.1 model with 405 billion parameters for all features are shown.

# 5. Results

## 5.1 Data Analysis

The performance of different LLMs in creating structured reports were evaluated and compared on the labeled cross-institutional datasets in terms of F1 score, 95% CI, and McNemar's test for each nodule feature, which measures both recall and precision. The average F1 score of all nodule features was used to measure the overall performance, the larger the better. To analyze statistics, z-test was used to compare different population proportions. While its accuracy is reflected by the above-evaluated LLM performance for structured reporting, the nodule feature retrieval system was also qualitatively evaluated.

## 5.2 Automatic Creation of Fully-Structured Reports

The F1 scores and 95% CI of different LLMs for creating FSR from free-text LSR are presented in Figures 6 and 7. On the average performance over all lung nodule features, our method consistently improved all the open-source LLMs, including LLAMA-3.1 8B by 18.42% (F1 score: 80.34% (95% CI: 77.91%-82.88%) vs 61.92% (95% CI: 58.74%-65.34%), LLAMA-3.1 70B by 5.34% (89.44% vs 84.10%), LLAMA-3.1 405B by 10.42% (97.60% vs 87.18%), Qwen-2 72B by 4.54% (87.50% vs 82.96%), Mistral-Large 123B by 7.07% (90.81% vs 83.74%), GPT-4o by 17.18% (97.60% vs 80.42%) on the Institution-1 dataset; and improved LLAMA-3.1 8B by 15.26% (80.70% vs 65.44%), LLAMA-3.1 70B by 6.65% (90.16% vs 83.51%), LLAMA-3.1 405B by 5.53% (96.92% vs 91.39%), Qwen-2 72B by 7.27% (86.84% vs 79.57%), Mistral-Large 123B by 3.89% (91.86% vs 87.97%) on the Institution-2 dataset.

For the best open-source LLAMA-3.1 405B model in extracting specific nodule features, our method significantly improved the F1 scores with p-value < 0.01 for all nodule features on the Institution-1 dataset, improved the F1 scores for 21 of 27 features and had the same F1 scores for the rest 8 features, among which 16 features had p-value < 0.01, on the Institution-2 dataset. Our method also improved the F1 scores of GPT-4o for 26 of 27 nodule features with 25 features having p-value < 0.01 on the Institution-1 dataset.

**Table 4. Distributions of nodule features on the Institution-1-Unlabeled dataset based on LLM-Created structured reports.**

| | | Age and Sex | | | | | | | |
|---|---|---|---|---|---|---|---|---|---|
| | | <55 y | | 55-65 y | | 65-75 y | | ≥75 y | |
| Features | Total | M | F | M | F | M | F | M | F |
| Total | 10377 (100) | 66 (0.6) | 120 (1.2) | 2164 (20.9) | 2263 (21.8) | 2317 (22.3) | 2535 (24.4) | 442 (4.3) | 470 (4.5) |
| Lobe | | | | | | | | | |
| LUL | 1687(16.3) | 9(13.6) | 22 (18.3) | 360 (16.6) | 361 (16.0) | 379 (16.4) | 404 (15.9) | 68 (15.4) | 84 (17.9) |
| Lingula | 210 (2.0) | 3 (4.5) | 3 (2.5) | 41(1.9) | 46 (2.0) | 41 (1.9) | 55 (2.2) | 8 (1.8) | 13 (2.8) |
| LLL | 1385 (13.3) | 12 (18.2) | 18 (15.0) | 303 (14.0) | 267 (11.8) | 315 (13.6) | 358 (14.1) | 57 (12.9) | 55 (11.7) |
| RUL | 2460 (23.7) | 19 (28.8) | 36 (30.0) | 502 (23.2) | 546 (24.1) | 533 (23.0) | 575 (22.7) | 123 (27.8) | 126 (26.8) |
| RML | 783 (7.5) | 2 (3.0) | 11 (9.2) | 192 (8.9) | 176 (7.8) | 167 (7.2) | 163 (6.4) | 43 (9.7) | 29 (6.2) |
| RLL | 1543 (14.9) | 3 (4.5) | 19 (15.8) | 286 (13.2) | 350 (15.5) | 326 (14.1) | 423 (16.7) | 68 (15.4) | 68 (14.5) |
| Lung | | | | | | | | | |
| Left | 3523 (34.0) | 27 (40.9) | 43 (35.8) | 748 (34.6) | 737 (32.6) | 811 (35.0) | 857 (33.8) | 141 (31.9) | 159 (33.8) |
| Right | 5194 (50.1) | 29 (43.9) | 67 (55.8) | 1071 (49.5) | 1162 (51.3) | 1131 (48.8) | 1262 (49.8) | 240 (54.3) | 232 (49.4) |
| Type | | | | | | | | | |
| Ground Glass Nonsolid | 696 (6.7) | 4 (6.1) | 13 (10.8) | 97 (4.5) | 186 (8.2) | 105 (4.5) | 200 (7.9) | 23 (5.2) | 68 (14.5) |
| Part-solid | 203 (2.0) | 1 (1.5) | 4 (3.3) | 40 (1.8) | 44 (1.9) | 35 (1.5) | 58 (2.3) | 12 (2.7) | 9 (1.9) |
| Solid | 2027 (19.5) | 28 (42.4) | 20 (16.7) | 439 (20.3) | 491 (21.7) | 400 (17.3) | 492 (19.4) | 94 (21.3) | 63 (13.4) |
| Status | | | | | | | | | |
| Decrease | 121 (1.2) | 2 (3.0) | 2 (1.7) | 30 (1.4) | 28 (1.2) | 19 (0.8) | 29 (1.1) | 7 (1.6) | 4 (0.9) |
| Increase | 229 (2.2) | 0 (0.0) | 2 (1.7) | 41 (1.9) | 37 (1.6) | 41 (1.8) | 76 (3.0) | 17 (3.8) | 15 (3.2) |
| Interval development | 97 (0.9) | 0 (0.0) | 1 (0.8) | 28 (1.3) | 12 (0.5) | 19 (0.8) | 27 (1.1) | 4 (0.9) | 6 (1.3) |
| new | 851 (8.2) | 12 (18.2) | 10 (8.3) | 154 (7.1) | 183 (8.1) | 200 (8.6) | 200 (7.9) | 37 (8.4) | 55 (11.7) |
| Resolved | 76 (0.7) | 0 (0.0) | 1 (0.8) | 18 (0.8) | 15 (0.7) | 21 (0.9) | 16 (0.6) | 2 (0.5) | 3 (0.6) |
| Average diameter | | | | | | | | | |
| <6 mm | 6004 (57.9) | 36 (54.5) | 80 (66.7) | 1279 (59.1) | 1347 (59.5) | 1247 (53.8) | 1526 (60.2) | 219 (49.5) | 270 (57.4) |
| 6-10 mm | 1226 (11.8) | 8 (12.2) | 11 (9.1) | 236 (10.9) | 219 (9.7) | 314 (13.5) | 317 (12.5) | 80 (18.1) | 41 (8.7) |
| 10-15 mm | 246 (2.4) | 4 (6.1) | 4 (3.3) | 32 (1.5) | 37 (1.6) | 75 (3.2) | 56 (2.2) | 18 (4.1) | 20 (4.3) |
| ≥ 15 mm | 104 (1.0) | 1 (1.5) | 1 (0.8) | 22 (1.0) | 19 (0.9) | 20 (0.9) | 28 (1.1) | 6 (1.3) | 7 (1.5) |
| Overall Lung-RADS | | | | | | | | | |
| 1 | 1162 (22.4) | 10 (27.0) | 21 (29.2) | 290 (25.8) | 283 (24.4) | 272 (23.0) | 227 (18.9) | 30 (14.6) | 29 (13.9) |
| 2 | 3427 (66.0) | 21 (56.8) | 44 (61.1) | 712 (63.5) | 768 (66.1) | 767 (64.7) | 836 (69.7) | 138 (67.3) | 141 (67.5) |
| 3 | 238 (4.6) | 0 (0.0) | 2 (2.8) | 48 (4.3) | 48 (4.1) | 57 (4.8) | 53 (4.4) | 12 (5.9) | 18 (8.6) |
| 4A | 167 (3.2) | 1 (2.7) | 1 (1.4) | 37 (3.3) | 30 (2.6) | 41 (3.5) | 43 (3.6) | 5 (2.4) | 9 (4.3) |
| 4B | 85 (1.6) | 0 (0.0) | 1 (1.4) | 16 (1.4) | 10 (0.9) | 20 (1.7) | 19 (1.6) | 12 (5.9) | 7 (3.3) |
| 4X | 18 (0.3) | 0 (0.0) | 0 (0.0) | 4 (0.4) | 4 (0.3) | 6 (0.5) | 2 (0.2) | 1 (0.5) | 1 (0.5) |

Note.— The number of nodules with the ratio (%) in parentheses are reported. M=Male, F=Female, LUL= left upper lobe, LLL=left lower lobe, RUL=right upper lobe, RML=right middle lobe, RLL=right lower lobe

**Table 4. Distributions of different attenuations over average diameter, location, and stability features on the Institution-1-Unlabeled dataset based on LLM-created structured reports.**

| | Avg Diameter (mm) | | | Location | | | | | | Stability | | | |
|---|---|---|---|---|---|---|---|---|---|---|---|---|---|
| | <6 | 6-10 | ≥10 | RUL | RML | RLL | LUL | LML | LLL | Stable | New | Grow | Decrease |
| Attenuation | | | | | | | | | | | | | |
| Solid /2027 | 1539 (75.9) | 305 (15.0) | 55 (2.7) | 529 (26.1) | 176 (8.7) | 405 (20.0) | 384 (18.9) | 38 (1.9) | 332 (16.4) | 1343 (66.3) | 212 (10.5) | 60 (3.0) | 33 (1.6) |
| PS /203 | 88 (43.3) | 59 (29.1) | 38 (18.7) | 72 (35.5) | 13 (6.4) | 23 (11.3) | 55 (27.1) | 6 (3.0) | 28 (13.8) | 86 (42.4) | 36 (17.7) | 44 (21.7) | 8 (3.9) |
| GG /696 | 335 (48.1) | 114 (16.4) | 81 (11.6) | 226 (32.5) | 42 (6.0) | 112 (16.1) | 164 (23.6) | 21 (3.0) | 59 (8.5) | 388 (55.7) | 88 (12.6) | 41 (5.9) | 45 (6.5) |

Note.— The number of nodules with the ratio (%) in parentheses are reported. LUL= left upper lobe, LLL=left lower lobe, RUL=right upper lobe, RML=right middle lobe, RLL=right lower lobe, PS=part-solid, GG=ground glass / nonsolid.

## 5.3 Automatic Large-Scale Nodule Feature Statistics

Our experiments demonstrated that nodule-level statistics can be automatically calculated from FSRs without need for expensive labeling of original LSRs. Table 4 summarizes the statistics of different nodule features stratified by age and sex. The statistical results show that the number of the extracted nodules in the upper lobes is significantly (p-value < 0.01) greater than that in other lobes, the number of the right lung is significantly (p-value < 0.01) greater than that on the left lung, female had significantly more (p-value < 0.01) ground glass nodules than male, the distribution Lung RADS scores over 0, 1, 2, 3, 4A, 4B, 4X are 1.0%, 22.4%, 66%, 4.6%, 3.2%, 1.6%, and 0.3%, respectively, where 0.8% reports did not provide lung RADS scores. The size (average diameter) distribution of lung nodules over <6 mm, 6~10 mm, 10~15 mm, ≥15 mm was 57.9%, 11.8%, 2.4%, and 1.0%, respectively, where 27% nodules did not have their sizes reported. Also, the results show that 1.2% nodules were decreased, 2.2% were increased and 0.9% were interval development, 8.2% were new nodules, and 0.7% were resolved. Table 4 summarized the distributions of average diameter, location, and stability conditioned on different nodule attenuations. All the automatic statistical results are closely consistent with the previous findings [21-24].

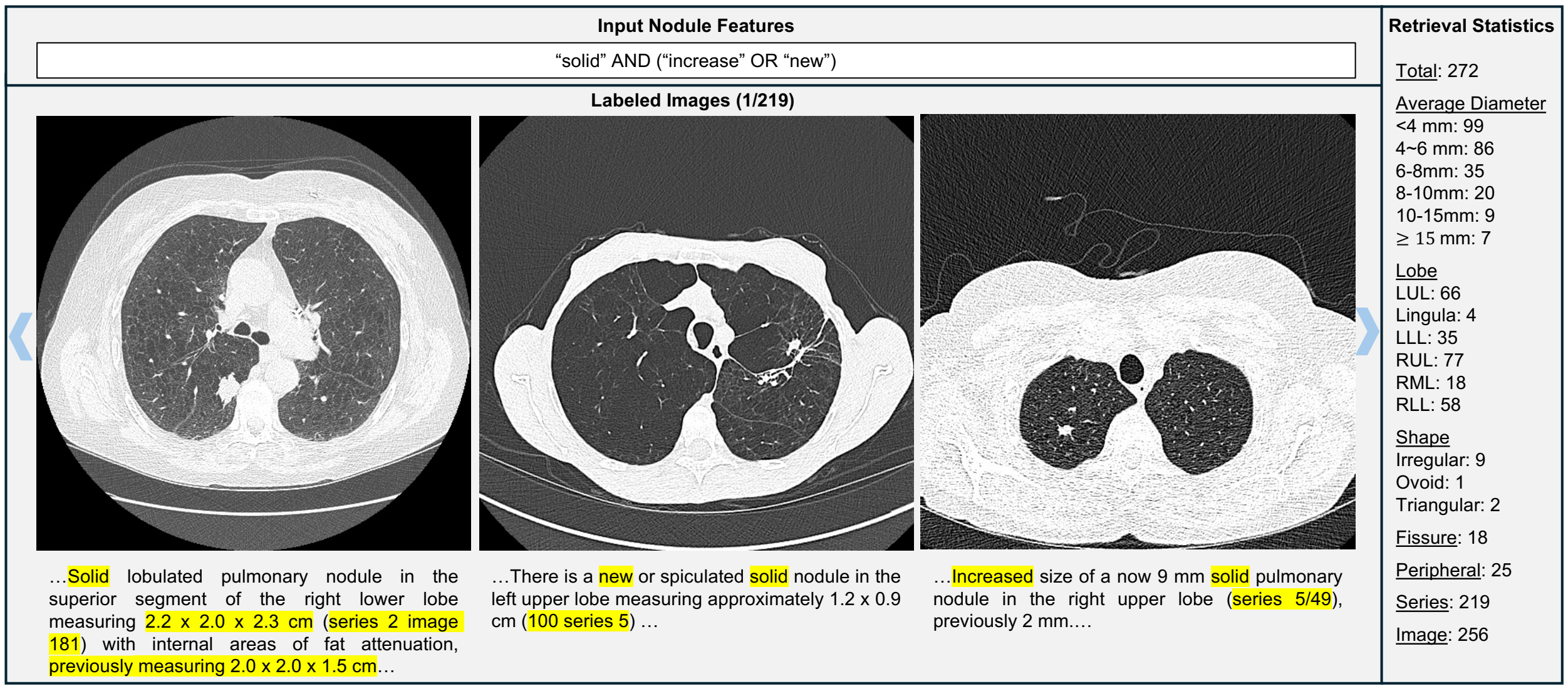

Figure 8. Advanced nodule-feature retrieval system based on structured radiology reports. By inputting *“solid AND (“increase” OR “new”)* in the Input Nodule Features section, 272 individual nodules met with the features were found, and the distributions over *average diameter, lobe, shape, fissure, and peripheral features* are shown in the Retrieval Statistics section. Among them, 219 nodules had the series number, and 256 nodules had the image number reported. The labeled images having both the series and image numbers are displayed (window level: -650, window width: 1500). Under each image, the original report is shown with the matched contents manually highlighted in yellow.

## 5.4 Complex Query Statistics and Individual Nodule Retrieval

Based on FSRs created by the best LLM, a nodule-level feature retrieval system was prototyped and tested on the 5192 LCS reports in the Institution-1-Unlabeled dataset. A retrieval example is shown in Figure 6, where given the combination of retrieval features *“solid” AND (“increase” OR*

*“new”)*, 272 individual nodules met with the features were found and the distributions over average diameter, lobe, shape, fissure, and peripheral features are calculated. Among them, 219 nodules had the series number, and 256 nodules had the image number. The labeled images having both the series and image numbers are displayed. Under each image, the original report is shown with the matched contents manually highlighted in yellow. In contrast to the keyword searching strategy in the original free-text reports, our retrieval system based on the created FSRs can capture semantic meanings. For the first example in Figure 8, our system correctly searched the increased nodule with the description “*measuring 2.0x2.0x2.3 cm…, previously measuring 2.0x2.0x1.5 cm…*”, although there is no “increase” mentioned.

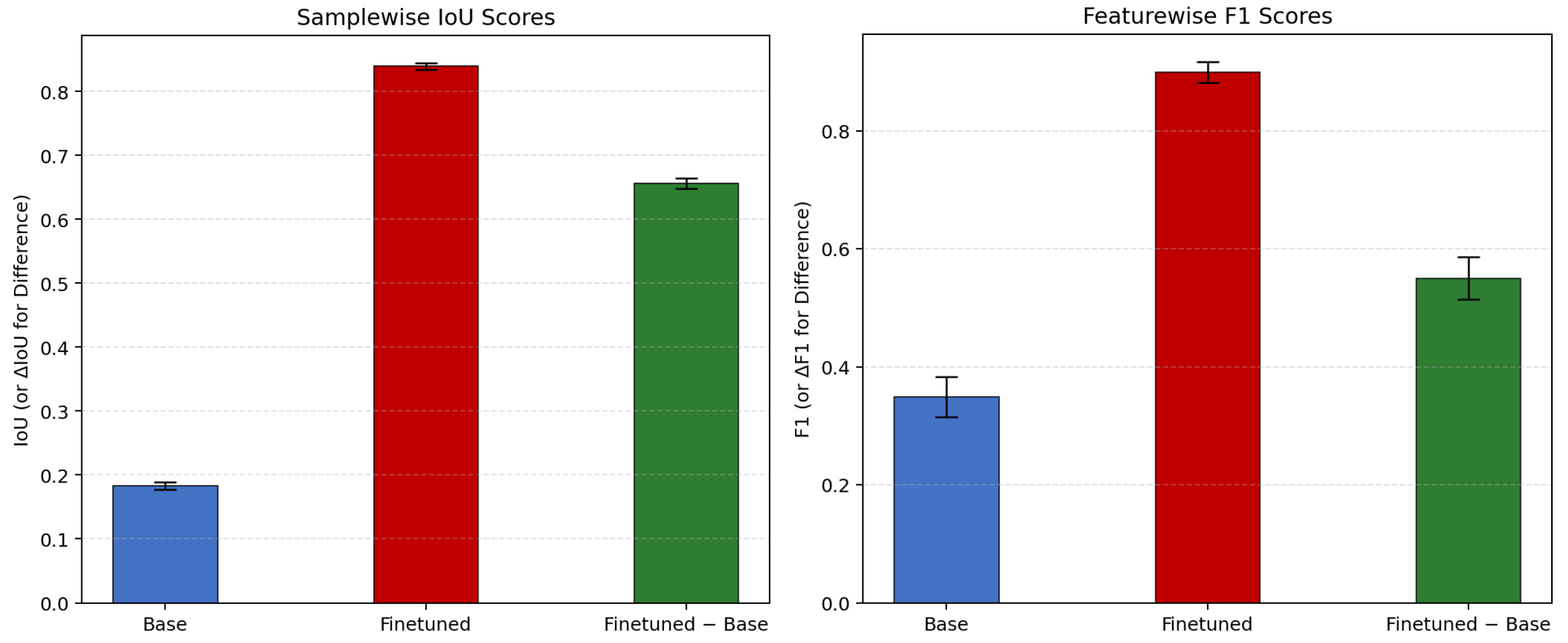

**Figure 9.** *Samplewise Intersection over Union (IoU) scores (left) and featurewise F1 scores for the lungs and pleura (right).* The base model showed limited structural alignment (mean IoU ≈ 0.18) and modest per-feature performance, while the finetuned model achieved high consistency (mean IoU ≈ 0.83) and near-perfect F1 scores across most pulmonary features, including effusion, emphysema, and nodule descriptors.

## 5.5 Evaluation of LLM Fine-tuning

The fine-tuned 8B model demonstrated markedly higher structural accuracy compared with the base model. On the 5,000-report evaluation set, sample-wise Intersection over Union (IoU) provided a global measure of alignment between predicted and reference JSON structures. As shown in Figure 9 (left), the base model achieved a mean±standard deviation IoU of only 0.1832±0.2043, indicating poor structural fidelity. In contrast, the finetuned model reached an IoU of 0.8396±0.1960, reflecting robust agreement across both categorical and numeric attributes. The relative improvement (ΔIoU ≈ +0.6564±0.2811) underscores the substantial benefit of domain-specific fine-tuning.

Complementary featurewise F1 analysis quantified how these improvements were distributed across individual features and anatomical regions. The aggregate featurewise F1 comparison (Figure 9, right) confirmed that the base model achieved a mean±standard deviation F1 of 0.3490±0.2446, whereas the fine-tuned model attained 0.8990±0.1284, yielding an improvement of approximately +0.5500±0.2542. On the full inference dataset of 50,000 samples, the fine-tuned model obtained an F1 score of 0.8963±0.1282, which is similar to the performance in the smaller dataset. This gain highlights that fine-tuning not only improved global JSON structure but also enhanced consistency in extracting clinically meaningful details across features.

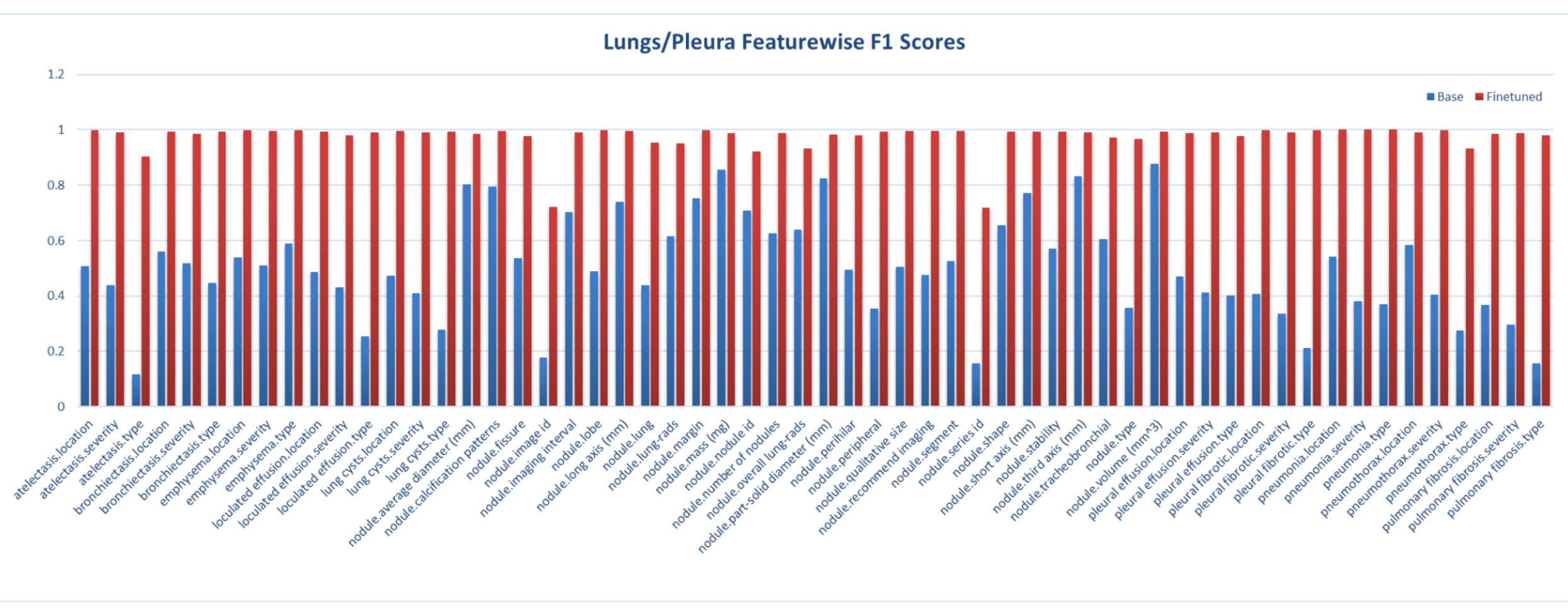

**Figure 10.** *Featurewise F1 scores for the lungs and pleura.* The base model achieved only modest performance (F1 ≈ 0.2–0.6) on common findings such as atelectasis, emphysema, and pleural effusion, and often failed to capture detailed attributes like emphysema severity, nodule margin, or pleural fibrosis. In contrast, the finetuned model consistently reached near-perfect F1 scores (≥0.95) across both common and rare features, demonstrating robust generalization and clinically meaningful improvements.

Region-level breakdowns further illustrate these improvements. In the lungs and pleura (Figure 10), the base model produced modest F1 scores ranging from 0.2 to 0.6 for common features such as atelectasis and effusion, and often failed on nuanced descriptors such as emphysema severity, nodule margin, and pleural fibrosis. The finetuned model, by contrast, consistently achieved near-perfect scores (0.95–1.0) across both common and rare features. In the mediastinum, hila, and axilla (Figure 11), the base model frequently failed to extract meaningful lymph node or mediastinal attributes, with F1 scores approaching zero. Finetuning elevated most features to >0.85, though residual variability persisted for rare attributes such as mediastinal shift or sub-region labeling. For the thoracic inlet and central airways (Figure 12), the base model again struggled, with F1 values generally <0.2. The fine-tuned model reached near-perfect scores across thyroid and airway descriptors, including nodule size and substernal extension. Similarly, in the heart and vessels (Figure 13), the base model rarely exceeded 0.3 on chamber abnormalities or vascular features. The finetuned model consistently achieved F1 scores in the 0.8–1.0 range, even for quantitative measurements such as vessel diameters and chamber sizes. Finally, in the chest wall and musculoskeletal (MSK) system (Figure 14), the base model demonstrated moderate reliability (0.4–0.6) for common features such as clavicle and rib size, but performed poorly on deformities, implants, and rare skeletal findings. The finetuned model corrected these deficits, delivering consistently high F1 scores (>0.9) across nearly all features.

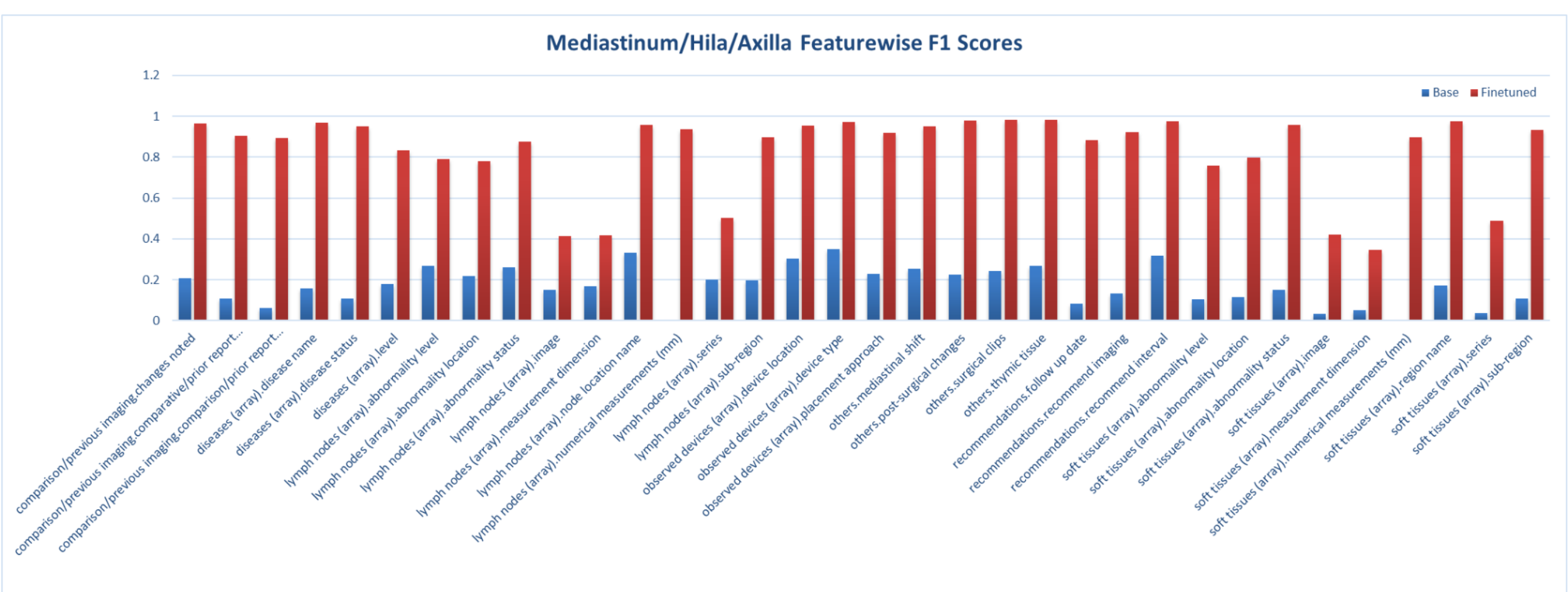

**Figure 11.** *Featurewise F1 scores for the mediastinum, hila, and axilla.* Finetuning substantially improved the extraction of lymph node attributes, mediastinal structures, and post-surgical findings, with most features achieving F1 > 0.85 compared with near-zero performance from the base model.

Taken together, the IoU and featurewise F1 analyses provide complementary perspectives. IoU captured overall structural alignment per sample, while F1 revealed fine-grained improvements across anatomical features. Both metrics demonstrate that finetuning transformed the model from one that often collapsed on complex or low-frequency features into one that achieved clinically reliable performance across diverse structures in radiology reports.

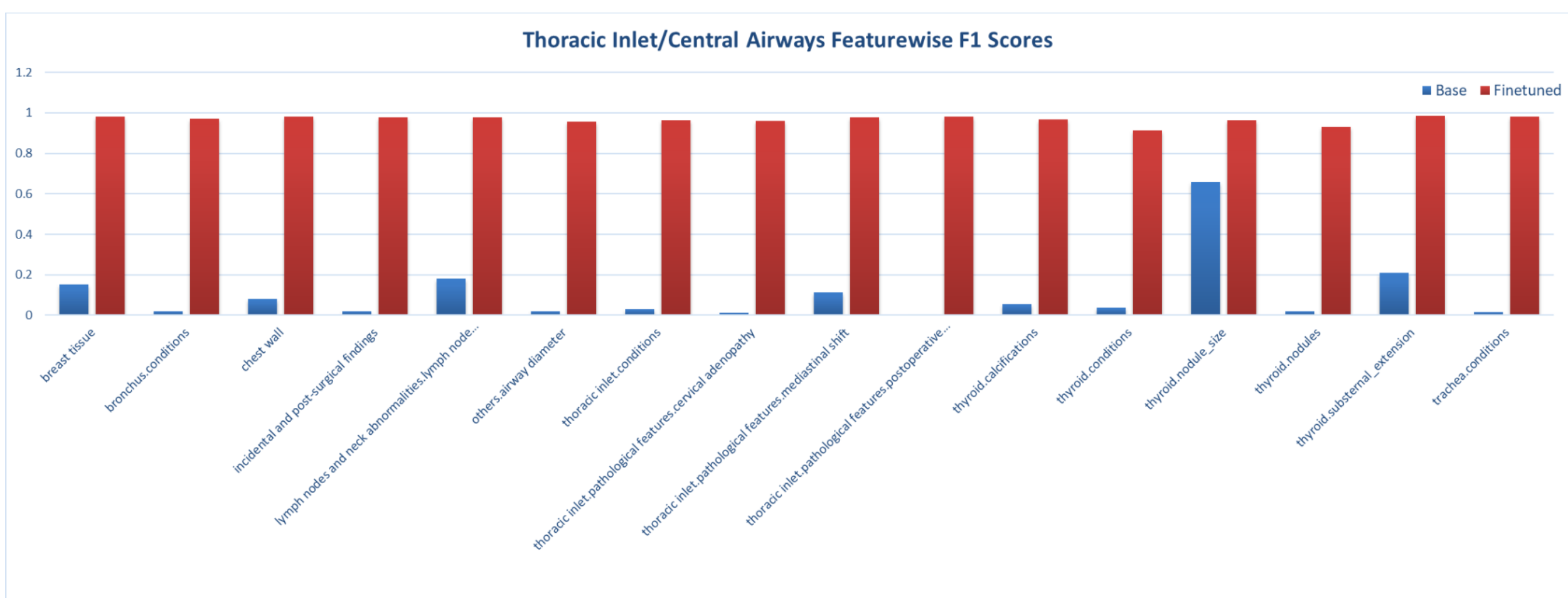

**Figure 12.** *Featurewise F1 scores for the thoracic inlet and central airways.* The base model failed to capture thyroid and airway descriptors reliably (F1 < 0.2), whereas the finetuned model achieved consistent accuracy across thyroid nodules, substernal extension, and tracheal conditions.

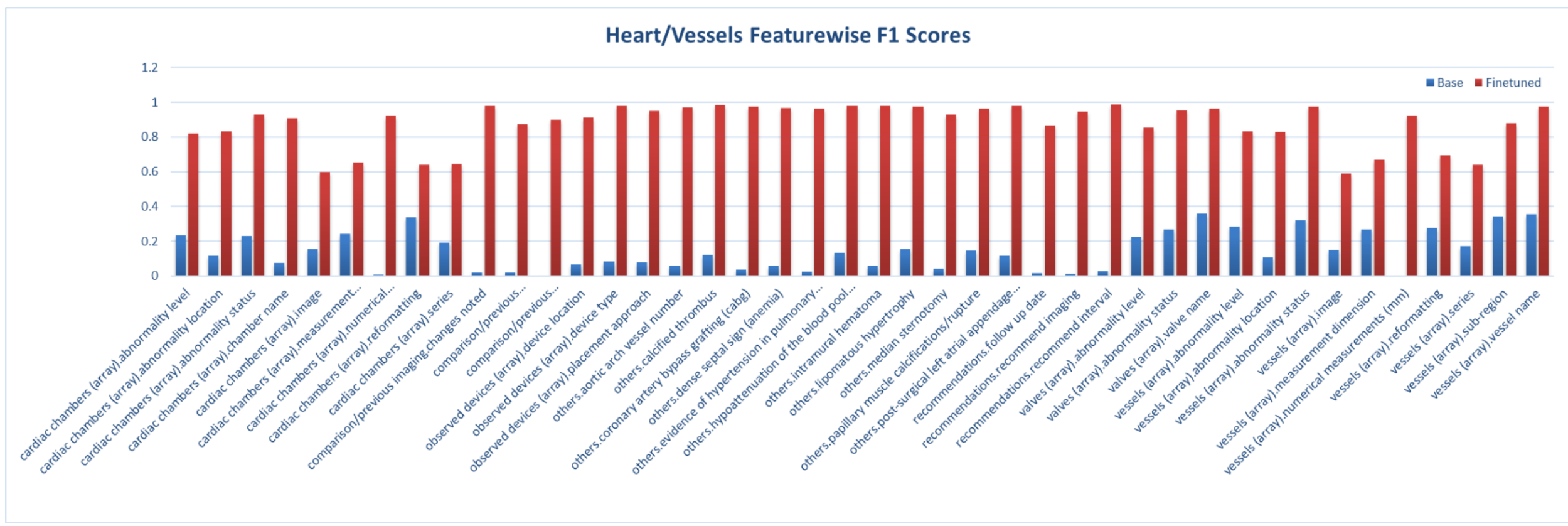

**Figure 13.** *Featurewise F1 scores for the heart and vessels.* Finetuning enabled accurate parsing of cardiac chamber abnormalities, vascular measurements, and surgical findings, with F1 scores approaching 1.0 across nearly all features, compared with <0.3 for the base model.

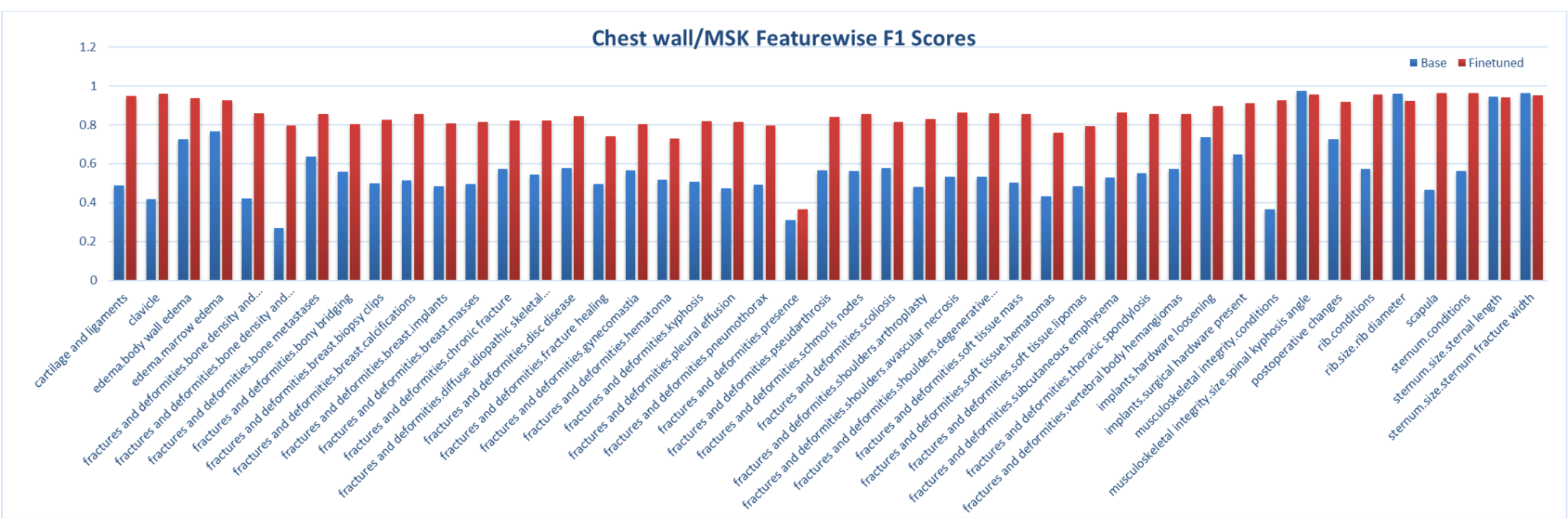

**Figure 14.** *Featurewise F1 scores for the chest wall and musculoskeletal (MSK) system*. The base model showed only moderate reliability on common skeletal features and failed on rarer abnormalities, while the finetuned model delivered consistent, near-perfect F1 scores across clavicle, ribs, sternum, and implant-related features.

# 6. Discussion

Fully-structured radiology reporting is desired to improve the quality, consistency, and actionability of radiology reports. However, FSR is rarely adopted due to lacking an efficient IT tool. This study developed and validated a dynamic-template-constrained LLM for accurately creating FSR without disrupting radiologists' attention and habits. Our enhanced LLM achieved a high accuracy with 97.60% and 96.92% F1 scores for creating FSR in LCS as evaluated on the cross-institutional datasets (n=500). Since the lung nodule reporting in this study is complex involving varying numbers of nodules and various nodule features, achieving such a high accuracy has strongly demonstrated the feasibility of our LLM-based tool for FSR.

Current LLMs struggle to generate predefined FSRs reliably, due to the absence of a robust mechanism for eliminating formatting errors and preventing content hallucinations. In contrast, our dynamic template-constrained encoding cannot only handle varying numbers of reported nodules with a dynamic template but also ensure zero formatting errors and zero content hallucinations. This is because our designed LLM inference leverages the template to strictly constrain LLM outputting correct template format tokens and selecting a predefined candidate value for each feature. Additionally, the template can be dynamically adjusted to accommodate varying numbers of reported nodules. As a result, our dynamic-template-constrained inference method improved the best LLM by 10.42% (97.60% vs 87.18%) and 5.53% (96.92% vs 91.39%) and improved the GPT-4o by 17.18% (97.60% vs 80.42%) on the cross-institutional datasets. On the other hand, most of the existing studies used proprietary LLMs, e.g., GPT-4o, which requires uploading reports to the cloud server with major privacy concerns. Because of the policy and cost, previous studies were conducted on relatively small datasets. Fortunately, the latest Llama 3.1 model with 405 billion parameters released on July 23, 2024, has a competitive performance with proprietary LLMs. Our developed software, vLLM-Structure, can locally deploy and consistently enhance this best open-source LLM with our inference approach, enabling large-scale studies without privacy issues.

Thanks to the standardized and quantified items in structured reporting, large-scale data mining becomes easy. We demonstrated two use cases of FSR in automatic statistical analysis and individual lung nodule retrieval. We conducted an automatic statistical analysis on a large-scale dataset (n=5192) that consists of 3-year consecutive LDCT LCS radiology reports in the real-world workflow. The automatically derived statistical distributions of attenuation, location, size, stability, and Lung-RADS were closely consistent with prior findings. The high evaluation accuracy and the consistent statistical results to prior findings suggest the effectiveness and reliability of LLM-based statistics. In our prototyped retrieval system, the inputs can be any combination of nodule features, then all nodules met with the input features will be listed individually and the distributions of the complex query features over various other features are calculated. The images containing the retrieved nodules having both the series and image numbers are displayed. In contrast, the current keyword-matching-based searching system cannot distinguish, visualize, or analyze individual nodules statistically. Moreover, our LLM-enabled retrieval system can capture the semantic meanings beyond the keyword matching from the free-text reports.

There are several potential clinical and research implications of our proposed dynamic-template-constrained LLM for FSR. By referencing a predefined template with rich, standardized features, LLM-generated FSRs can assist with real-time quality checks and alerts during interpretation, helping radiologists avoid incomplete report information. In LCS, these FSRs can present key information from prior reports, enabling follow-up CT reports to focus on the stability and changes in lung nodule characteristics. Automatic conversion of LSR to standardized FSR with the LLM can also help make radiology reports consistent across institutions. The discrete, standardized features of LLM-generated FSRs make large-scale data mining and multi-institutional statistics on historical reports straightforward, facilitating real-world surveillance of various abnormality characteristics for optimized patient management. The quantitative data from these structured reports can also be directly used to train AI models and validate AI-derived findings on large-scale datasets [25, 26]. Our LLM-based nodule-level retrieval system supports complex statistical analyses, flexible individual searches, and visualizations, which can be used for diagnosis, research, teaching, and other purposes. Furthermore, linking structured radiology reports with more electronic medical records can help build more advanced systems. The structured and standardized content of LLM-generated FSRs may also enable the accurate generation of automatic impressions [27], improving reporting efficiency.

This study had several limitations. We conducted only retrospective experiments, generating FSRs from existing LSRs. Future prospective studies could deploy our software in clinical workflows to evaluate the effectiveness of creating structured reports through interactions between LLMs and radiologists. Additionally, we developed standardized templates and created structured reports specifically for lung nodule reporting, without addressing other potential abnormalities. However, as lung nodule reporting is a core component of LCS, it serves as a representative and extendable framework. Our future work will focus on addressing the remaining findings in LDCT LCS and extending our LLM solution to other radiology reports beyond LCS and CT. While we created a standardized template for lung nodules in this cross-institutional study, it has not yet been formally adopted. Nevertheless, our findings are informative and encouraging, demonstrating the potential of our LLM-based approach for fully-structured radiology reporting.

This study also demonstrates that smaller LLMs, when carefully finetuned on synthetic medical reports, can serve as efficient and accurate parsers for converting free-text radiology reports into structured feature tables. Unlike large foundation models, which require extensive prompting and incur significant computational overhead, the finetuned lightweight model internalized the reporting schema and operated within the context window of 8B-class models. This enabled faster inference, reduced token usage, and eliminated truncation errors, making large-scale deployment feasible.

In conclusion, we have proposed a dynamic template-constrained LLM to FSRs from the free-text LSR. Our method has achieved the state-of-the-art performance on cross-institutional datasets, without formatting errors, content hallucinations, or privacy issues. We have successfully demonstrated the utilities and merits of structured reporting in automatic statistical analysis and advanced lung nodule retrieval on a large-scale dataset. Our open-source software, vLLM-Structure, is publicly available for further translational research.

## References

[1] J. M. Nobel, K. van Geel, and S. G. F. Robben, "Structured reporting in radiology: a systematic review to explore its potential," *European Radiology,* vol. 32, no. 4, pp. 2837-2854, 2022/04/01, 2022.

[2] D. B. Larson, A. J. Towbin, R. M. Pryor, and L. F. Donnelly, "Improving consistency in radiology reporting through the use of department-wide standardized structured reporting," *Radiology,* vol. 267, no. 1, pp. 240-250, 2013.

[3] T. Jorg, M. C. Halfmann, G. Arnhold, D. Pinto dos Santos, R. Kloeckner, C. Düber, P. Mildenberger, F. Jungmann, and L. Müller, "Implementation of structured reporting in clinical routine: a review of 7 years of institutional experience," *Insights into Imaging,* vol. 14, no. 1, pp. 61, 2023/04/11, 2023.

[4] D. Leithner, E. Sala, E. Neri, H.-P. Schlemmer, M. D'Anastasi, M. Weber, G. Avesani, I. Caglic, D. Caruso, M. Gabelloni, V. Goh, V. Granata, W. G. Kunz, S. Nougaret, L. Russo, R. Woitek, and M. E. Mayerhoefer, "Perceptions of radiologists on structured reporting for cancer imaging—a survey by the European Society of Oncologic Imaging (ESOI)," *European Radiology,* vol. 34, no. 8, pp. 5120-5130, 2024/08/01, 2024.

[5] J. M. Nobel, E. M. Kok, and S. G. F. Robben, "Redefining the structure of structured reporting in radiology," *Insights into Imaging,* vol. 11, no. 1, pp. 10, 2020/02/04, 2020.

[6] R. European Society of, "ESR paper on structured reporting in radiology," *Insights into Imaging,* vol. 9, no. 1, pp. 1-7, 2018/02/01, 2018.

[7] D. P. dos Santos, E. Kotter, P. Mildenberger, L. Martí-Bonmatí, and R. European Society of, "ESR paper on structured reporting in radiology—update 2023," *Insights into Imaging,* vol. 14, no. 1, pp. 199, 2023/11/23, 2023.

[8] D. Harris, D. M. Yousem, E. A. Krupinski, and M. Motaghi, "Eye-tracking differences between free text and template radiology reports: a pilot study," *Journal of Medical Imaging,* vol. 10, no. S1, pp. S11902-S11902, 2023.

[9] J. Achiam, S. Adler, S. Agarwal, L. Ahmad, I. Akkaya, F. L. Aleman, D. Almeida, J. Altenschmidt, S. Altman, and S. Anadkat, "Gpt-4 technical report," *arXiv preprint arXiv:2303.08774*, 2023.

[10] L. C. Adams, D. Truhn, F. Busch, A. Kader, S. M. Niehues, M. R. Makowski, and K. K. Bressem, "Leveraging GPT-4 for Post Hoc Transformation of Free-text Radiology Reports into Structured Reporting: A Multilingual Feasibility Study," *Radiology,* vol. 307, no. 4, pp. e230725, 2023.

[11] R. Bhayana, B. Nanda, T. Dehkharghanian, Y. Deng, N. Bhambra, G. Elias, D. Datta, A. Kambadakone, C. G. Shwaartz, C.-A. Moulton, D. Henault, S. Gallinger, S. Krishna, and K. Fowler, "Large Language Models for Automated Synoptic Reports and Resectability Categorization in Pancreatic Cancer," *Radiology,* vol. 311, no. 3, pp. e233117, 2024.

[12] Q. Lyu, J. Tan, M. E. Zapadka, J. Ponnatapura, C. Niu, K. J. Myers, G. Wang, and C. T. Whitlow, "Translating radiology reports into plain language using ChatGPT and GPT-4 with prompt learning: results, limitations, and potential," *Visual Computing for Industry, Biomedicine, and Art,* vol. 6, no. 1, pp. 9, 2023/05/18, 2023.

[13] H. Jiang, S. Xia, Y. Yang, J. Xu, Q. Hua, Z. Mei, Y. Hou, M. Wei, L. Lai, N. Li, Y. Dong, and J. Zhou, "Transforming free-text radiology reports into structured reports using ChatGPT: A study on thyroid ultrasonography," *European Journal of Radiology,* vol. 175, pp. 111458, 2024/06/01/, 2024.

[14] L. C. Adams, D. Truhn, F. Busch, F. Dorfner, J. Nawabi, M. R. Makowski, K. K. Bressem, and L. Moy, "Llama 3 Challenges Proprietary State-of-the-Art Large Language Models in Radiology Board–style Examination Questions," *Radiology,* vol. 312, no. 2, pp. e241191, 2024.

[15] A. Dubey, A. Jauhri, A. Pandey, A. Kadian, A. Al-Dahle, A. Letman, A. Mathur, A. Schelten, A. Yang, and A. Fan, "The llama 3 herd of models," *arXiv preprint arXiv:2407.21783*, 2024.

[16] T. Nakaura, R. Ito, D. Ueda, T. Nozaki, Y. Fushimi, Y. Matsui, M. Yanagawa, A. Yamada, T. Tsuboyama, N. Fujima, F. Tatsugami, K. Hirata, S. Fujita, K. Kamagata, T. Fujioka, M. Kawamura, and S. Naganawa, "The impact of large language models on radiology: a guide for radiologists on the latest innovations in AI," *Japanese Journal of Radiology,* vol. 42, no. 7, pp. 685-696, 2024/07/01, 2024.

[17] Y. Shen, L. Heacock, J. Elias, K. D. Hentel, B. Reig, G. Shih, and L. Moy, "ChatGPT and Other Large Language Models Are Double-edged Swords," *Radiology,* vol. 307, no. 2, pp. e230163, 2023.

[18] W. Kwon, Z. Li, S. Zhuang, Y. Sheng, L. Zheng, C. H. Yu, J. Gonzalez, H. Zhang, and I. Stoica, "Efficient memory management for large language model serving with pagedattention." pp. 611-626.

[19] A. Yang, B. Yang, B. Hui, B. Zheng, B. Yu, C. Zhou, C. Li, C. Li, D. Liu, and F. Huang, "Qwen2 technical report," *arXiv preprint arXiv:2407.10671*, 2024.

[20] Mistral. "https://huggingface.co/mistralai/Mistral-Large-Instruct-2407," Sep 23, 2024.

[21] J. E. Walter, M. A. Heuvelmans, and M. Oudkerk, "Small pulmonary nodules in baseline and incidence screening rounds of low-dose CT lung cancer screening," *Translational Lung Cancer Research,* vol. 6, no. 1, pp. 42-51, 2017.

[22] A. R. Larici, A. Farchione, P. Franchi, M. Ciliberto, G. Cicchetti, L. Calandriello, A. Del Ciello, and L. Bonomo, "Lung nodules: size still matters," *European respiratory review,* vol. 26, no. 146, 2017.

[23] J. Christensen, A. E. Prosper, C. C. Wu, J. Chung, E. Lee, B. Elicker, A. R. Hunsaker, M. Petranovic, K. L. Sandler, B. Stiles, P. Mazzone, D. Yankelevitz, D. Aberle, C. Chiles, and E. Kazerooni, "ACR Lung-RADS v2022: Assessment Categories and Management Recommendations," *Journal of the American College of Radiology,* vol. 21, no. 3, pp. 473-488, 2024.

[24] J. Cai, M. Vonder, G. J. Pelgrim, M. Rook, G. Kramer, H. J. M. Groen, G. H. de Bock, R. Vliegenthart, and A. de Roos, "Distribution of Solid Lung Nodules Presence and Size by Age and Sex in a Northern European Nonsmoking Population," *Radiology,* vol. 312, no. 2, pp. e231436, 2024.

[25] C. Niu, Q. Lyu, C. D. Carothers, P. Kaviani, J. Tan, P. Yan, M. K. Kalra, C. T. Whitlow, and G. Wang, "Specialty-Oriented Generalist Medical AI for Chest CT Screening," *arXiv:2304.02649*, 2024.

[26] C. Niu, and G. Wang, "Unsupervised contrastive learning based transformer for lung nodule detection," *Physics in Medicine & Biology,* vol. 67, no. 20, pp. 204001, 2022.

[27] L. Zhang, M. Liu, L. Wang, Y. Zhang, X. Xu, Z. Pan, Y. Feng, J. Zhao, L. Zhang, G. Yao, X. Chen, X. Xie, L. Moy, and S. Atzen, "Constructing a Large Language Model to Generate Impressions from Findings in Radiology Reports," *Radiology,* vol. 312, no. 3, pp. e240885, 2024.

## APPENDIX: A Dynamic-Template-Constrained Large Language Model

To eliminate the formatting errors and content hallucinations of an LLM, we propose dynamic-template-constrained decoding for inference. The basic idea is to use a structured and standardized template to strictly constrain the LLM output. As shown in Figure 2, our LLM inference method takes the system instruction and the conventional LSR as input, uses the structured and standardized template to strictly constrain every output token during decoding, and then produces the FSR accordingly. Our approach is in sharp contrast to the existing studies, which focus on designing the system instruction to prompt LLMs. However, such prompting cannot eliminate either formatting errors or content hallucinations.

Specifically, we designed a dynamic template-constrained decoding method shown in Figure 3. In the common LLM inference process in Figure 3 (a), the concatenation of system instructions and free-text reports are first tokenized, the causal Transformer autoregressively calculates the probability distribution over all tokens, then a sampling strategy is implemented to sample the next token, and finally the sampled tokens are converted to the output text, which is also appended to the input token sequence for the next iteration. Our dynamic template-constrained inference is shown in Figure 3 (b), with the differences from the common inference highlighted in orange.

Before explaining the differences, let us introduce how to deal with the predefined template in Table 1. We represent the structured and standardized template in a dynamic JSON format with three kinds of text: template format text without highlight, special template text highlighted in pink, and candidate text highlighted in green, as shown in Figure 3 (c) and (d), respectively. Every special template text has a candidate set defined in Table 1. To handle varying numbers of nodules reported in different reports, a dynamic template is designed to dynamically determine and create the required number of nodule descriptors (a single nodule descriptor consists of multiple features in the black box of Figure 3 (c)) during inference.

In our dynamic template-constrained decoding, the concatenation of a system instruction, a free-text report, and a template is first tokenized. Here we added the special template text into the tokenizer so that they can be recognized and localized in the template. There are three kinds of template tokens including: (1) template format tokens converted from the template format text, (2) special template tokens converted from special template text, and (3) candidate tokens converted from candidate text. As shown in Figure 3 (c) and (d), the next token is autoregressively predicted along the order of dynamic template. During the next token prediction, we deterministically choose the token with the maximum probability among its candidate tokens. The candidate tokens of the template format token only contain itself so that the output is exactly the predefined format text, ensuring no JSON formatting errors. When a special template token is recognized, its predefined candidate token set will be used. For the example in Figure 3 (d) to predict the tokens for “|<lobe>|”, there are seven predefined candidates, and each candidate text has been tokenized into a sequence. All the seven candidate sequences are aligned from scratch. In predicting “right lower lobe”, the first token is selected among “right”, “left”, “ling” and “null” with the maximum probability. After the “right” token is chosen, all other candidate sequences without “right” will be ignored in predicting the subsequent tokens. This process will be repeated until a complete candidate text is decoded. The number of nodule descriptors will be predicted first to adjust the template by creating the required number of nodule descriptors. When the predicted number of nodule descriptors is zero, the “nodules” part in the JSON template will be removed.